\documentclass[acmsmall]{acmart}

\AtBeginDocument{%
  }

\usepackage{multirow}
\usepackage{multicol}
\usepackage{makecell}
\usepackage{color,soul}
\usepackage{caption}
\usepackage{subcaption}
\usepackage{threeparttable}
\usepackage[table]{xcolor} 

\definecolor{mypink}{RGB}{192, 0, 0}
\definecolor{myorange}{RGB}{235, 108, 21}
\definecolor{myblue}{RGB}{68, 114, 196}

\DeclareCaptionFormat{mybluecaption}{%
  \textcolor{blue}{#1#2}
  #3
}
\begin{document}

\setcopyright{acmlicensed}
\acmJournal{CSUR}
\acmDOI{10.1145/3770916}

\title{A Survey of Defenses against AI-generated Visual Media: Detection, Disruption, and Authentication}

\author{Jingyi Deng}
\affiliation{%
  \institution{School of Cyber Science and Engineering, Xi'an Jiaotong University}
  \city{Xi'an}
  \state{Shaanxi}
  \country{China}}
\email{misscc320@stu.xjtu.edu.cn}

\author{Chenhao Lin}
\authornote{Corresponding author.}
\authornotemark[0]
\affiliation{%
  \institution{School of Cyber Science and Engineering, Xi'an Jiaotong University and Interdisciplinary Research Center of Frontier science and technology, Xi'an Jiaotong University}
  \city{Xi'an}
  \state{Shaanxi}
  \country{China}}
\email{linchenhao@xjtu.edu.cn}

\author{Zhengyu Zhao}
\affiliation{%
  \institution{School of Cyber Science and Engineering, Xi'an Jiaotong University}
  \city{Xi'an}
  \state{Shaanxi}
  \country{China}}
\email{zhengyu.zhao@xjtu.edu.cn}

\author{Shuai Liu}
\affiliation{%
 \institution{School of Software Engineering, Xi'an Jiaotong University}
  \city{Xi'an}
  \state{Shaanxi}
  \country{China}}
\email{sh\_liu@mail.xjtu.edu.cn}

\author{Zhe Peng}
\affiliation{%
 \institution{Department of Industrial and Systems Engineering, Hong Kong Polytechnic University}
  \city{Hong Kong}
  \state{Hong Kong}
  \country{China}}
\email{jeffrey-zhe.peng@polyu.edu.hk}

\author{Qian Wang}
\affiliation{%
 \institution{School of Cyber Science and Engineering, Wuhan University}
  \city{Wuhan}
  \state{Hubei}
  \country{China}}
\email{qianwang@whu.edu.cn}

\author{Chao Shen}
\affiliation{%
 \institution{School of Cyber Science and Engineering, Xi'an Jiaotong University}
  \city{Xi'an}
  \state{Shaanxi}
  \country{China}}
\email{chaoshen@mail.xjtu.edu.cn}

\renewcommand{\shortauthors}{Deng et al.}

\begin{abstract}
  Deep generative models have demonstrated impressive performance in various computer vision applications, including image synthesis, video generation, and medical analysis. Despite their significant advancements, these models may be used for malicious purposes, such as misinformation, deception, and copyright violation. In this paper, we provide a systematic and timely review of research efforts on defenses against AI-generated visual media, covering detection, disruption, and authentication. We review existing methods and summarize the mainstream defense-related tasks within a unified passive and proactive framework. Moreover, we survey the derivative tasks concerning the trustworthiness of defenses, such as their robustness and fairness. For each defense strategy, we formulate its general pipeline and propose a multidimensional taxonomy applicable across defense tasks, based on methodological strategies. Additionally, we summarize the commonly used evaluation datasets, criteria, and metrics. Finally, by analyzing the reviewed studies, we provide insights into current research challenges and suggest possible directions for future research.
\end{abstract}


\received{02 September 2024}
\received[revised]{10 May 2025}
\received[accepted]{24 July 2025}

\maketitle

\section{Introduction}\label{sec:introduction}
The authenticity of digital vision media has become increasingly challenging due to advancements in deep generative models, given their remarkable capabilities in image and video synthesis and editing. These deep generative models primarily include variational autoencoders (VAEs)~\cite{kingma2013auto}, generative adversarial networks (GANs)~\cite{goodfellow2014generative}, flow models~\cite{rezende2015variational}, autoregressive models~\cite{esser2021taming}, diffusion models (DMs)~\cite{rombach2022high, ruiz2023dreambooth}, and fractal generative models~\cite{li2025fractal}. Initially, constrained by their limited generative ability to generate diverse content~\cite{zhang2019self, brock2018large}, most generative models focused on single-class generation. This leads to the development of a manipulation technique known as Deepfake~\cite{FaceSwap, karras2020analyzing, karras2019style, nirkin2019fsgan}, which primarily generates visual content containing human faces. More recently, impressive advancements, particularly in DMs, have enabled the generation of visual media depicting complex objects and scenes with high fidelity and diversity~\cite{esser2021taming, rombach2022high}. These breakthroughs have been applied to various computer vision tasks, including image and video generation~\cite{singer2022make}, image super-resolution~\cite{saharia2022image}, and anomaly detection~\cite{wyatt2022anoddpm}.

While generative content offers positive impacts, it also poses risks of technology abuse~\cite{chen2023pathway}, such as misinformation~\cite{xu2023combating}, deception~\cite{mustak2023deepfakes}, and copyright violations~\cite{somepalli2023diffusion}, raising ethical concerns to society~\cite{hagendorff2024mapping, mustak2023deepfakes}. Additionally, the proliferation of generative media on the Internet leads to data contamination~\cite{hataya2023will, ravuri2019classification}, posing new challenges for the research community. Consequently, ensuring media authenticity and identity traceability is crucial to mitigate the potential misuse, prompting numerous explorations.

One popular way to mitigate the threat of generative media is to (passively) detect the authenticity of a given visual media. Early approaches primarily aimed at identifying Deepfake~\cite{zhou2017two, li2018ictu, li2018exposing} and GAN-generated~\cite{mccloskey2018detecting, nataraj2019detecting} media, typically relying on hand-crafted artifact features. However, due to the lack of acknowledged benchmarking datasets, these methods were difficult to compare directly. Since 2019, various large-scale datasets have been  proposed~\cite{rossler2019faceforensics++, li2020celeb, dolhansky2020deepfake, dang2020detection}, providing manipulated data with larger scale, improved quality, and greater diversity. Accordingly, reliance on hand-crafted features has diminished, with deep learning-based discriminative representations being increasingly exploited~\cite{li2020face, dang2020detection}. During these advances, most detection approaches have demonstrated strong in-domain performance but limited generalization across domains. As a result, generalization ability has emerged as a key evaluation criterion. In parallel, with the growing diversity of generative models, detection methods now require to generalize to a broader range of types of generative media. Beyond generalizability, robustness has become another critical evaluation criterion, as visual media shared on real-world social platforms is often subject to lossy compression and degradation. Throughout this evolution, fine-grained detection tasks (e.g., forgery localization, forgery attribution) and trustworthiness of detection have also been explored, collaboratively contributing to the advancement of detection technologies.

Another effective way to prevent the malicious use of generative media is to (proactively) modify the generative media before they are released.
On the one hand, one can directly disrupt the potentially generated outputs by adversarially perturbing the inputs to generative models~\cite{ruiz2020disrupting, huang2021initiative}, thereby preventing their generation and potential misuse. On the other hand, digital watermarks can be injected into the original or generative media to embed identification information~\cite{yu2021artificial} or highlight forgery traces~\cite{asnani2023malp} for further media verification. This provides traceability of generative media and mitigates misuse. Additionally, the trustworthiness of proactive defenses is also explored~\cite{jiang2023evading, zhao2024can}.

With the growing interest in both passive and proactive defenses against AI-generated visual media, several existing surveys~\cite{tolosana2020deepfakes, mirsky2021creation, yu2021survey, juefei2022countering, seow2022comprehensive, rana2022deepfake, cardenuto2023age, lin2024detecting, tariang2024synthetic, wang2023security} have aimed to analyze and categorize existing works. 
Nevertheless, these surveys have several limitations that hinder their ability to provide systematic and comprehensive summaries. 
First, they either lack coverage of proactive defense strategies such as authentication and disruption~\cite{tolosana2020deepfakes, mirsky2021creation, yu2021survey, juefei2022countering, seow2022comprehensive, rana2022deepfake, chen2023challenges, wang2023security, cardenuto2023age, lin2024detecting, tariang2024synthetic}, or they consider a limited set of detection tasks~\cite{wang2023security, chen2023challenges}.
Second, they lack a multidimensional method taxonomy that is commonly applicable across various defense-related tasks. For instance, while Lin et al.~\cite{lin2024detecting} and Xu et al.~\cite{juefei2022countering} propose taxonomies on detection, their taxonomies are only based on a single perspective and does not apply to the detection tasks such as forgery localization.
Third, they do not discuss the trustworthiness and evaluation methodologies of various defenses from a systematic perspective.
Specifically, although Xu et al.~\cite{juefei2022countering} and Lin et al.~\cite{lin2024detecting} have discussed the evasion of detection, they lack a comprehensive discussion of trustworthiness concerns.
Detailed comparisons between existing surveys and ours are summarized in Table~\ref{tab:related_surveys}.

\begin{table}
\caption{Comparisons between existing related surveys and ours. Ours discusses disruption and authentication defense strategies in addition to detection, each with a proposed multidimensional taxonomy applicable across defense tasks. Ours also covers a more complete review of defense trustworthiness and evaluation methodology.
\label{tab:related_surveys}}
\centering
\resizebox{1\textwidth}{!}{
\begin{tabular}{l|c|cc|cc|cc|cc|cccc}
\hline
\multirow{2}{*}{Surveys} &
\multirow{2}{*}{Year} &
\multicolumn{2}{c|}{Detection} & 
\multicolumn{2}{c|}{Disruption} & 
\multicolumn{2}{c|}{Authentication} &
\multicolumn{2}{c|}{Method Taxonomy} &
\multicolumn{4}{c}{Evaluation Methodology}\\
&& Task & Trust & Task & Trust & Task & Trust & Dims & Universal & Methods & Datasets & Criteria & Metrics\\
\hline
Tolosana et al.~\cite{tolosana2020deepfakes} & 2020 & 1 & -- & -- & -- & -- & -- & -- & -- & \checkmark & \checkmark & -- & \checkmark \\
Misky et al.~\cite{mirsky2021creation} & 2021 & 1 & 1 & -- & -- & -- & -- & 1 & -- & \checkmark & -- & -- & --\\
Xu et al.~\cite{juefei2022countering} & 2022 & 2 & 1 & -- & -- & -- & -- & 1 & -- & \checkmark &\checkmark & -- & \checkmark\\
Seow et al.~\cite{seow2022comprehensive} & 2022 & 1 & -- & -- & -- & -- & -- & 1 & -- & \checkmark & \checkmark & -- & \checkmark \\
Cardenuto et al.~\cite{cardenuto2023age} & 2023 & 1 & 1 & -- & -- & -- & -- & 1 & -- & \checkmark & -- & -- & -- \\
Chen et al.~\cite{chen2023challenges} & 2023 & -- & -- & 1 & -- & 1 & -- & 1 & -- & \checkmark & -- & -- & -- \\
Wang et al.~\cite{wang2023security} & 2023 & 2 & -- & 2 & -- & 1 & -- & 1 & -- & \checkmark & -- & -- & --\\
Lin et al.~\cite{lin2024detecting} & 2024 & 3 & 1 & -- & -- & -- & -- & 1 & -- & \checkmark & -- & -- & --\\
Tariang et al.~\cite{tariang2024synthetic} & 2024 & 2 & -- & -- & -- & -- & -- & 1 & -- & \checkmark & \checkmark & \checkmark & --\\
Ours & 2024 & 4 & 3 & 2 & 1 & 3 & 1 & 3 & \checkmark & \checkmark & \checkmark & \checkmark & \checkmark\\
\hline
\end{tabular}
}
\begin{threeparttable}
\begin{tablenotes}
    \tiny
      \item[*] The numbers in the table represent either the number of tasks covered in each paper or the number of trustworthiness topics considered.
      \item[*] The dimensions of a method taxonomy refer to the number of perspectives of the methodological strategy on which the taxonomy is based. A universal method taxonomy suggests that it is uniformly applicable across defense-related tasks.
\end{tablenotes}
\end{threeparttable}
\end{table}

\begin{figure}
\begin{center}
    \includegraphics[width=0.7\textwidth]{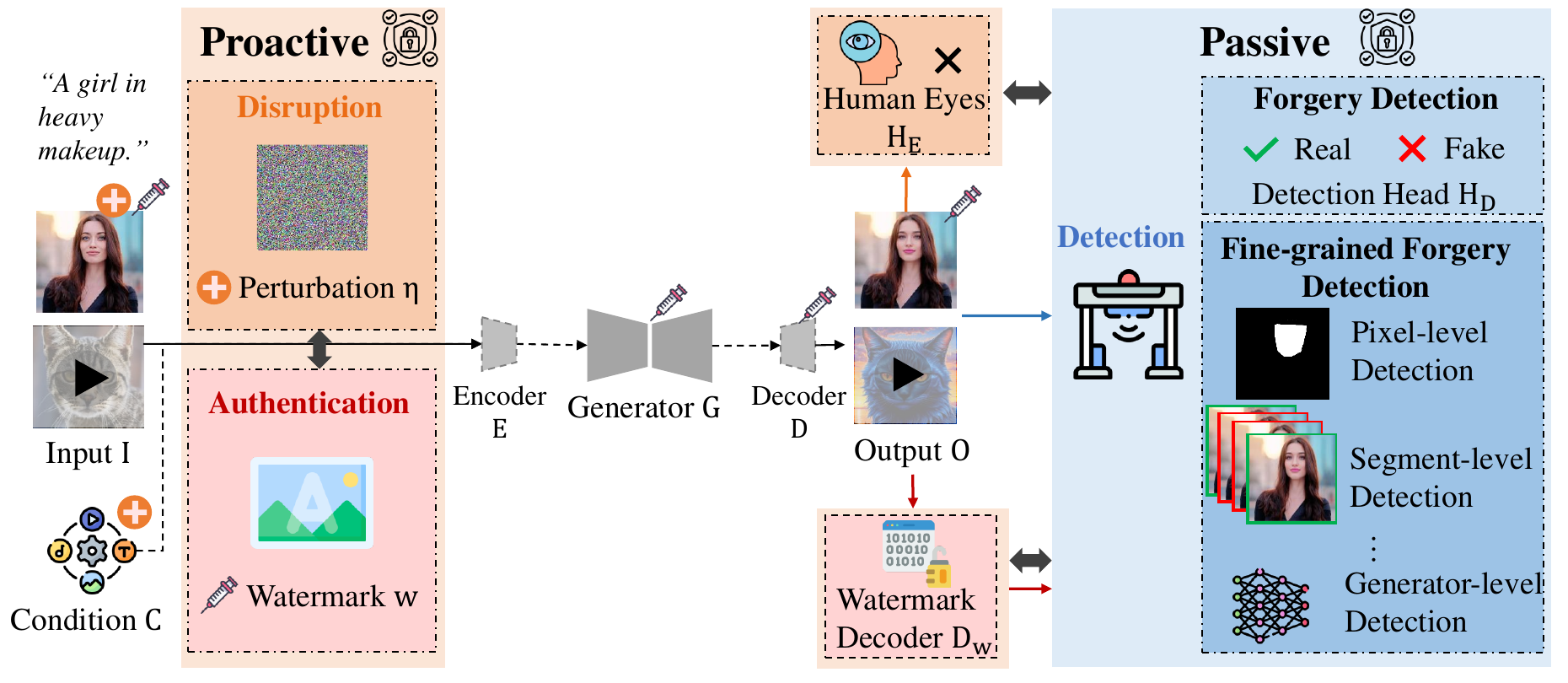}
\end{center}
   \caption{Framework of proactive and passive defenses against AI-generated visual media, which consists of three defense strategies: \textcolor{myorange}{disruption}, \textcolor{mypink}{authentication}, and \textcolor{myblue}{detection}. These three defense strategies can be employed independently or collaboratively. Within each independent defense strategy, we review related mainstream tasks and their trustworthiness research. In addition, we survey the research efforts on their joint defense.}
\label{fig:framework_overview}
\end{figure}

This paper aims to address these limitations by providing a systematic and comprehensive review of the up-to-date literature. Our main contributions are summarized as follows:

\begin{itemize}
\item We provide the first systematic and comprehensive review of various defenses against diverse AI-generated media in computer vision. It covers both passive and proactive defenses (i.e., detection, disruption, authentication) and their trustworthiness, within a unified framework following the generation pipeline.

\item For each defense strategy, we construct a common methodological pipeline to illustrate its involved tasks and trustworthiness concerns. Moreover, we develop a novel taxonomy for each defense strategy, applicable across defense tasks, based on commonly employed methodological strategies, and accordingly categorize the reviewed methods.

\item We provide an overview of common evaluation methodology for detection, disruption, and authentication, including datasets, criteria, and metrics.

\item We summarize the ongoing challenges in current research and present our outlook on potential areas and directions for future research.
\end{itemize}

The remainder of this paper is organized as follows: Section~\ref{sec:background} provides a brief background on defenses against AI-generated visual media, including targeted generation techniques and definition/formulation of the involved defense strategies; Section~\ref{sec:detection} summarizes detection and its trustworthiness; Section~\ref{sec:disruption} summarizes disruption and its trustworthiness; Section~\ref{sec:authentication} summarizes authentication and its trustworthiness; Section~\ref{sec:evluation_methodology} reviews the evaluation methodology, covering datasets, criteria, and metrics; Section~\ref{sec:recommendations_for_future_work} discusses the recommendations for future work; and Section~\ref{sec:conclusion} concludes with a discussion.

\section{Background}\label{sec:background}
This section presents background on defenses against AI-generated visual media. We begin by briefly introducing the generation techniques that most defense approaches target. We then define and formulate detection, disruption, and authentication, along with their trustworthiness, within a unified framework covering both passive and proactive strategies, following the common generation pipeline, as shown in Fig.~\ref{fig:framework_overview}.

\subsection{Generation of AI-generated Visual Media}\label{sec:Generation}
Deep generative models aim to capture the probability distribution of given data and generate new, similar samples. 

Early applications of VAEs~\cite{FaceSwap, FakeApp} and the GAN-based models~\cite{karras2019style, karras2020analyzing} primarily focused on single-class generation~\cite{epstein2023online}, particularly facial content manipulation, leading to the prevalence of \textit{deepfake}. Most existing deepfake defense methods are designed to detect fake facial content generated by four main techniques: entire face synthesis\cite{karras2020analyzing}, attribute manipulation\cite{karras2019style}, identity swap\cite{FaceSwap}, and face reenactment\cite{nirkin2019fsgan}. With continued advances in deep generative models, \textit{universal generative visual media} featuring complex scenes can now be synthesized and edited with high quality and diversity. Such visual media can be generated by a variety of deep generative models, including GANs~\cite{goodfellow2014generative}, flow models~\cite{rezende2015variational}, autoregressive models~\cite{esser2021taming}, DMs~\cite{rombach2022high, ruiz2023dreambooth}, and fractal generative models~\cite{li2025fractal}. Users can access this generative content through multiple sources, such as the Internet~\cite{kolomeets2024face, wei2024understanding}, third-party datasets~\cite{rossler2019faceforensics++, khalid2021fakeavceleb, ojha2023towards}, and self-generated data. Therefore, defending against AI-generated content has become increasingly important.

In this paper, we model a general generation pipeline that most defense methods target, as shown in Fig.~\ref{fig:framework_overview}. This pipeline includes an optional encoding process performed by encoders $E$ to encode inputs and conditions into embeddings in the latent space, a generation process carried out by generators $G$ (e.g., VAEs, GANs, and DMs), and an optional decoding process implemented by decoders $D$ to decode the generative embeddings into generative media.

\begin{figure}
\begin{center}
   \includegraphics[width=0.5\textwidth]{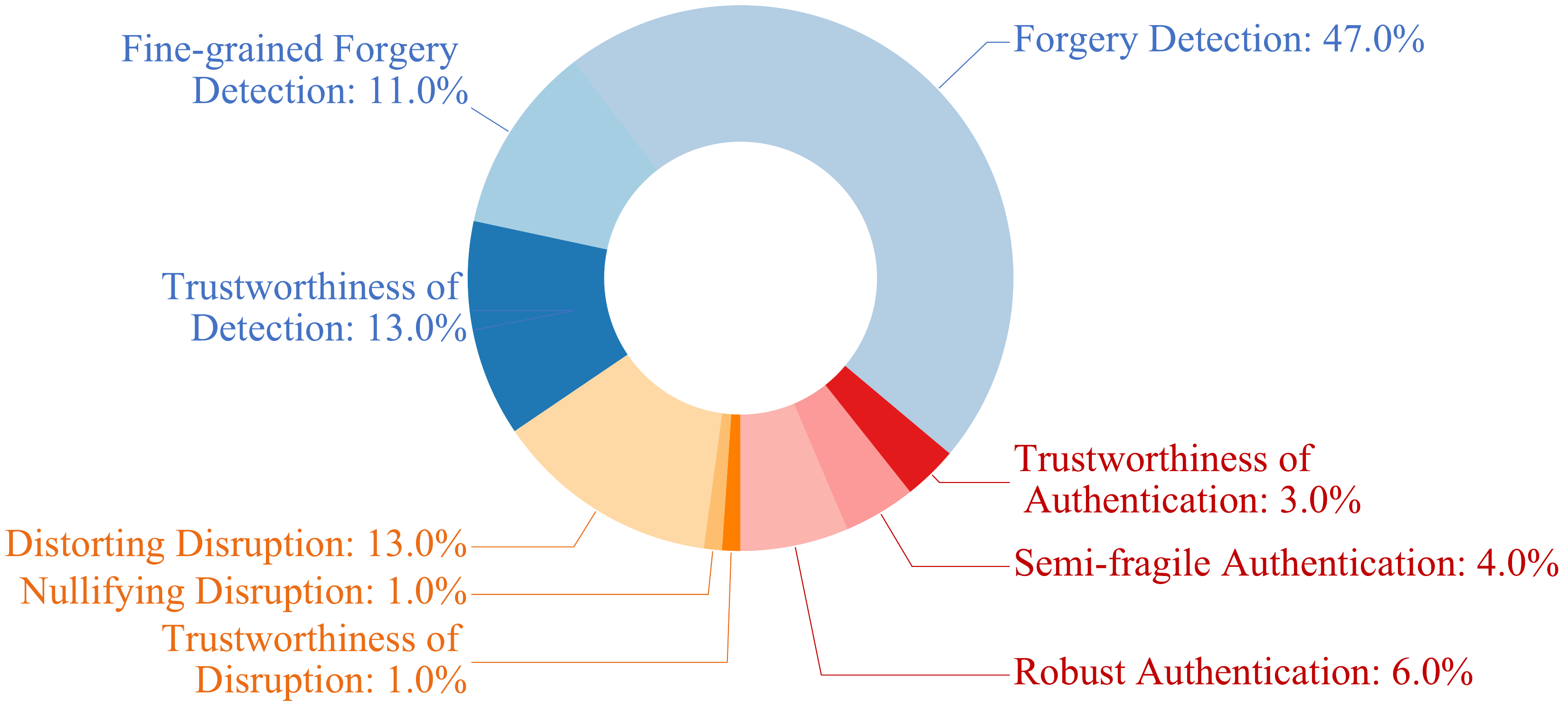}
\end{center}
   \caption{Distribution of literature across different defense strategies, including detailed statistics on the number of publications for the contained tasks and the defense associated trustworthiness: \textcolor{myorange}{disruption} (29 papers), \textcolor{mypink}{authentication} (26 papers), and \textcolor{myblue}{detection} (132 papers).}
\label{fig:defense_task_pie}
\label{fig:onecol}
\end{figure}

\subsection{Definition and Formulation}
Preventing the misuse of AI-generated visual media can be achieved by proactive and passive defenses, specifically via detection, disruption, and authentication, as illustrated in Fig.~\ref{fig:framework_overview}. Detection serves as a passive defense strategy that identifies forgeries after the generated media have been released, whereas disruption and authentication are proactive defense strategies that either disrupt or watermark the media before their release. Furthermore, given that the defense systems equipped by these three defense strategies are predominantly AI-based, they inherently raise trustworthiness concerns\cite{kemmerzell2025towards}, with robustness and fairness being the most critical requirements in this field. The literature distribution of these defense strategies, detailed with their involved mainstream tasks, and their trustworthiness are illustrated in Fig.~\ref{fig:defense_task_pie}. 

\subsubsection{Detection}
Detection identifies whether a visual media is generated by a generative model, i.e., forgery detection. If the media is found to be forged, detection can further perform fine-grained forgery detection (e.g., passive forgery localization, forgery attribution, sequential manipulation prediction).
Forgery detection is a coarse-grained binary classification task that predicts a media-level fakeness score using the detection head $H_D$. Fine-grained forgery detection predicts detailed forgery information with different detection heads.

In addition to detection itself, key concerns regarding detection trustworthiness include detection robustness and detection fairness.

\subsubsection{Disruption}\label{sec:disruption_formulation}
Disruption interferes with the generation process by guiding generative models to generate perceptually abnormal or unaltered media. This is achieved by adding small perturbations $\eta$ to the inputs or conditions of generation, resulting in perturbed inputs $I_{adv}$ or conditions $C_{adv}$. The generative media, $O_{adv}=G(I_{adv}, C_{adv})$, is then perceived as distorted or nullified by human eyes $H_E$. 

Beyond disruption itself, its trustworthiness concern of robustness is threatened by adversarial purification, which aims to eliminate the perturbations and thereby leads to ineffective disruption.



\subsubsection{Authentication}\label{sec:authentication_formulation}
Authentication involves injecting watermarks into AI-generated visual media before release, followed by decoders and detectors to verify the watermarks for specific purposes such as copyright protection, forgery authentication, and forgery localization. Specifically, a watermark $w$ can be embedded before, during, or after the generation process to produce a watermarked generative media $O_w$. For verification, a decoder $D_W$ decodes the estimated watermark $\hat{w}$ from $O_w$, denoted as $\hat{w} = D_W(O_w)$. Based on $w$ and $\hat{w}$, different authentication tasks can be performed using various detection heads.

Beyond authentication itself, robustness is a major trustworthiness concern, as watermarks are vulnerable to corruption, which may lead to evasion of authentication.

Overall, the three defense strategies outlined above can be independently applied to counter AI-generated visual media, as is common in existing defense methods. However, these strategies are not mutually exclusive, and recent studies have begun to explore both the positive~\cite{zhu2024watermark} and negative~\cite{wu2024watermarks, wang2022deepfake} effects of their integration.
\begin{figure}
\begin{center}
\begin{subfigure}[b]{0.4\textwidth}
    \includegraphics[width=\linewidth]{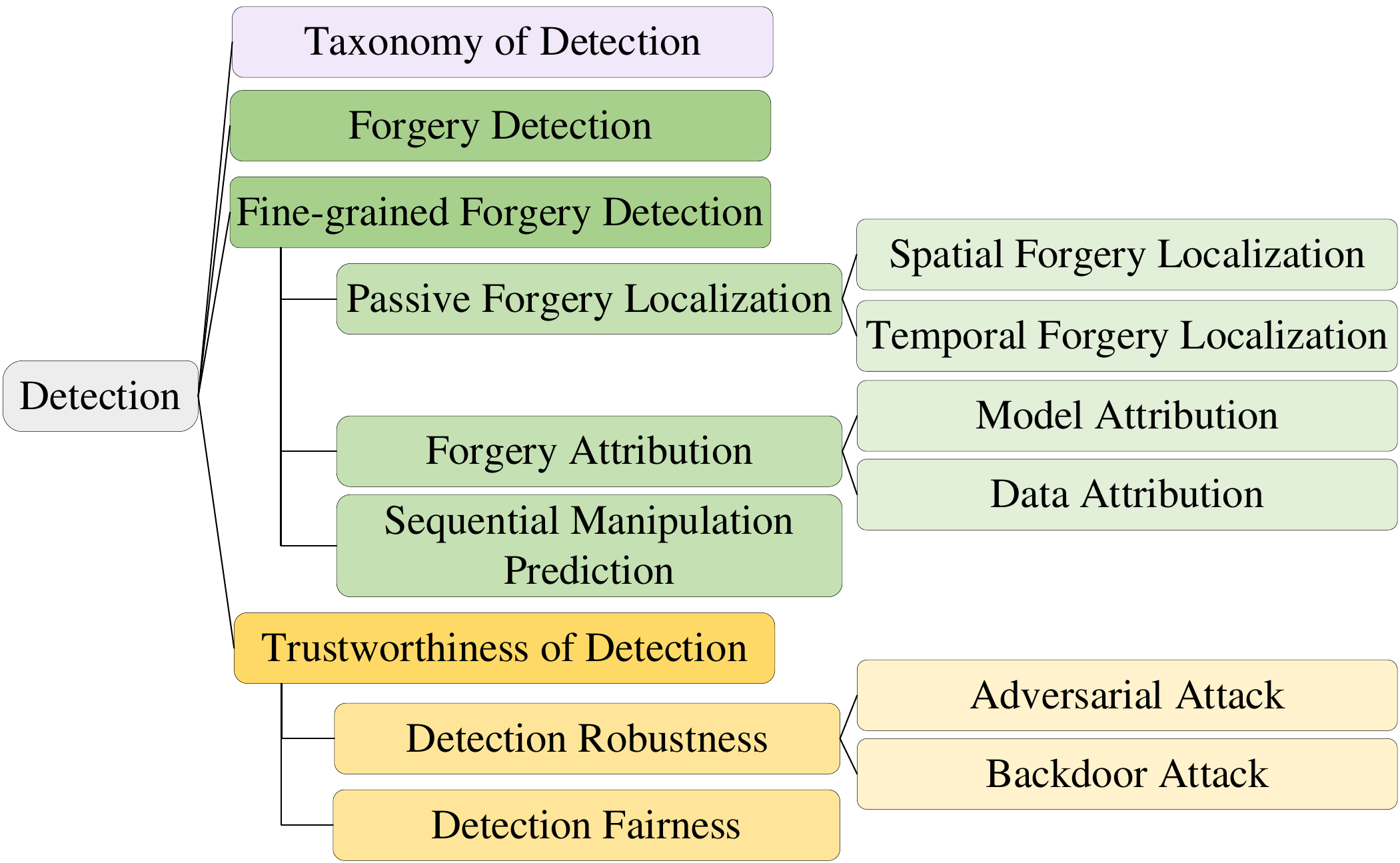}
    \caption{Structure of our paper on detection.}
    \label{fig:detection_paper_structure}
\end{subfigure}
\begin{subfigure}[b]{0.5\textwidth}
    \includegraphics[width=\linewidth]{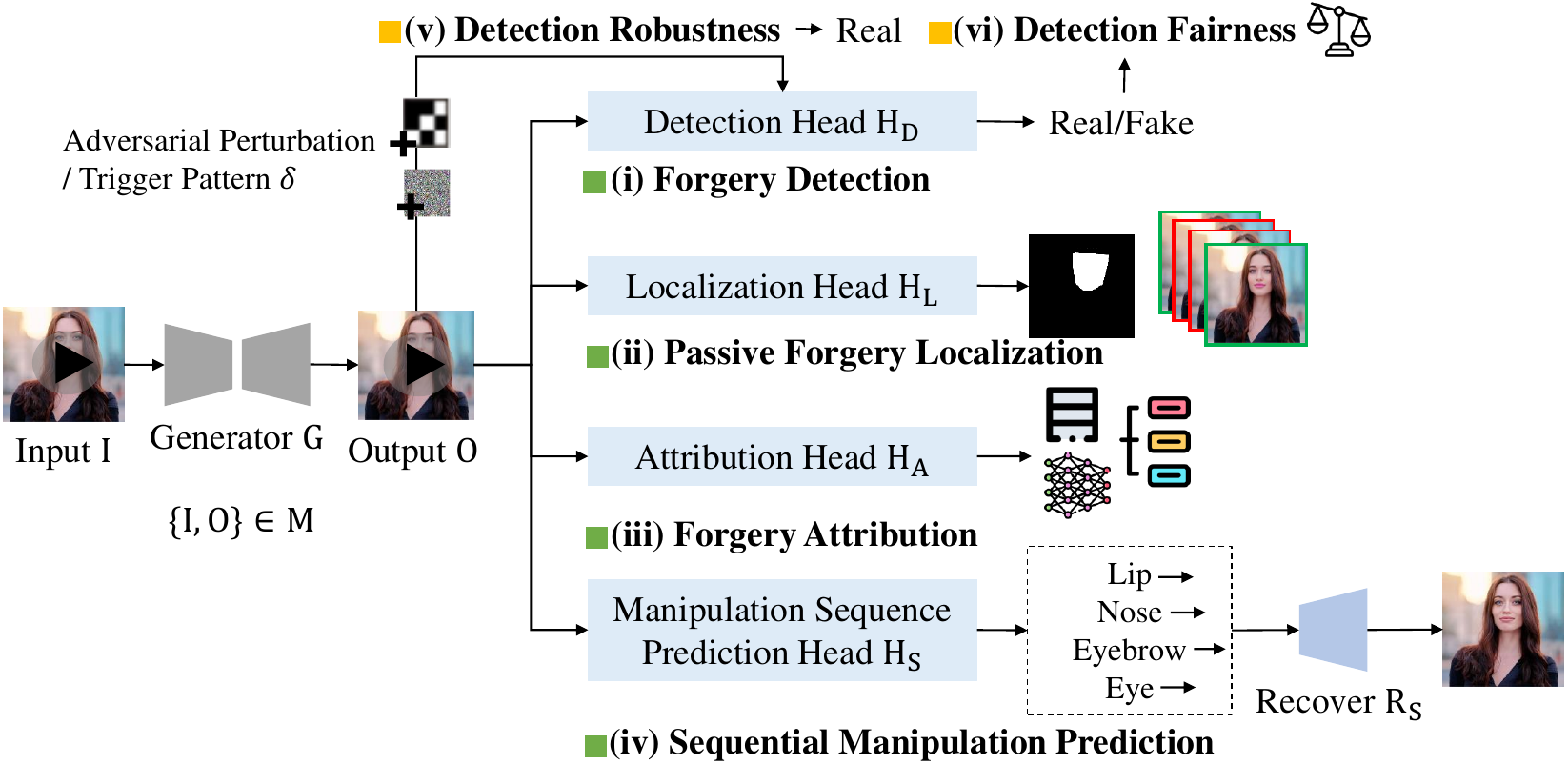}
    \caption{Overview of detection.}
    \label{fig:detection_overview}
\end{subfigure}
\end{center}
   \caption{Paper structure and overview of detection. Mainstream detection tasks ($\vcenter{\hbox{\includegraphics[width=8pt]{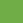}}}$) include (i) forgery detection (Section \ref{sec:forgery_detection}) and fine-grained forgery detection (e.g., (ii) passive forgery localization (Section \ref{sec:passive_forgery_localization}), (iii) forgery attribution (Section. \ref{sec:forgery_attribution}), (iv) sequential manipulation prediction (Section \ref{sec:sequential_manipulation_prediction})). Mainstream trustworthiness issues ($\vcenter{\hbox{\includegraphics[width=8pt]{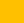}}}$) of detection include (v) detection robustness (Section \ref{sec:detection_robustness}) and (vi) detection fairness (Section \ref{sec:detection_fairness}).}
\end{figure}

\section{Detection}\label{sec:detection}
Detection includes both forgery detection and fine-grained forgery detection tasks. As illustrated in Fig.~\ref{fig:detection_paper_structure}, this section begins with the taxonomy of detection methods, followed by an overview of methods for each task based on the proposed taxonomy. Finally, we review detection trustworthiness. An overview of detection tasks and their trustworthiness is summarized in Fig.~\ref{fig:detection_overview}.

\subsection{Taxonomy of Detection}\label{sec:taxonomy_of_detection}
Detection methods identify various generative artifacts based on discriminative features learned by different representation learning strategies. Therefore, we classify existing detection methods from two perspectives: detected generative artifacts and adopted discriminative representation learning strategies. 

\subsubsection{Generative Artifacts}\label{sec:generative_artifacts}
Based on the modality of explored generative artifacts, generation artifacts can be categorized into image-level and video-level artifacts. Each category includes specific artifacts that can be explored either exclusively or jointly for detection.

\paragraph{Image-level Generative Artifacts}
Image-level artifacts refer to static forgery traces present in a generative image or a single frame image extracted from generative videos. This section summarizes the most commonly detected image-level generative artifacts.

\begin{itemize}
\item Physical Characteristics-based Artifacts. Physical characteristics, such as facial features, fingerprints, and irises, are widely used in biometrics for individual identification~\cite{sundararajan2018deep}. AI-generated media often exhibit abnormal physical characteristics due to the inherent challenges in generating natural ones.

\item Generation Pipeline-based Artifacts. Deep generative models and their generation processes often share common operations, such as the image blending operation~\cite{li2020face, shiohara2022detecting}, up-sampling, convolution, and batch normalization~\cite{durall2020watch, liu2021spatial, tan2024rethinking}. These common operations tend to leave artifacts that generalize across multiple types of synthesized content.

\item Semantic Artifacts. Generative media contains increasingly complex semantic contents, which can introduce anomalous semantic artifacts~\cite{bui2022repmix, sha2023fake, zheng2024breaking} that can be used for detection.

\item Spatial Artifacts. Generative images typically exhibit multi-scale spatial artifacts, ranging from low-level cues~\cite{chai2020makes, zhao2021multi} (e.g., corners, edges, and textures) to high-level artifacts~\cite{zhao2021multi} (e.g., objects, scenes, and their parts), which can be explored directly in RGB images. In this taxonomy, we define spatial artifacts as those present in the spatial domain but not classified as other artifacts.

\item Frequency Artifacts. 
The spectral distribution of the mid- and high- frequency bands differs between real and fake images: fake spectra usually contain more energy and exhibit various periodic patterns in the mid-to-high frequency bands, while real spectra decay gradually from low- to high- frequency bands~\cite{wang2020cnn, jeong2022frepgan, corvi2023intriguing}. This difference mainly arises from the common use of up-sampling operations in generation models and the absence of high-frequency noise in fake images, where the noise is typically introduced by the sensor or circuitry of a digital camera~\cite{fei2022learning, luo2021generalizing}. Furthermore, frequency artifacts have also been observed in the low-frequency band of deepfake images~\cite{zhou2024freqblender}.
Although such high-frequency artifacts exist, they can be easily degraded by common data augmentation techniques~\cite{wang2020cnn} and spectral correction operations~\cite{dong2022think, jeong2022frepgan}, which should be carefully considered when leveraging frequency artifacts.

\end{itemize}

\paragraph{Video-level Generative Artifacts}
Cross-frame consistency issues are common in AI-generated videos~\cite{huang2023vbench}. As a result, video-level generative artifacts, reflecting temporal discontinuities between frames, have been extensively studied. 
Spatial-temporal inconsistencies at varying scales, alongside multimodality artifacts, are typically explored. 
This section introduces video-level specific artifacts that can be leveraged individually or in combination to guide detection.

\begin{figure}[t]
\begin{center}
   \includegraphics[width=0.5\linewidth]{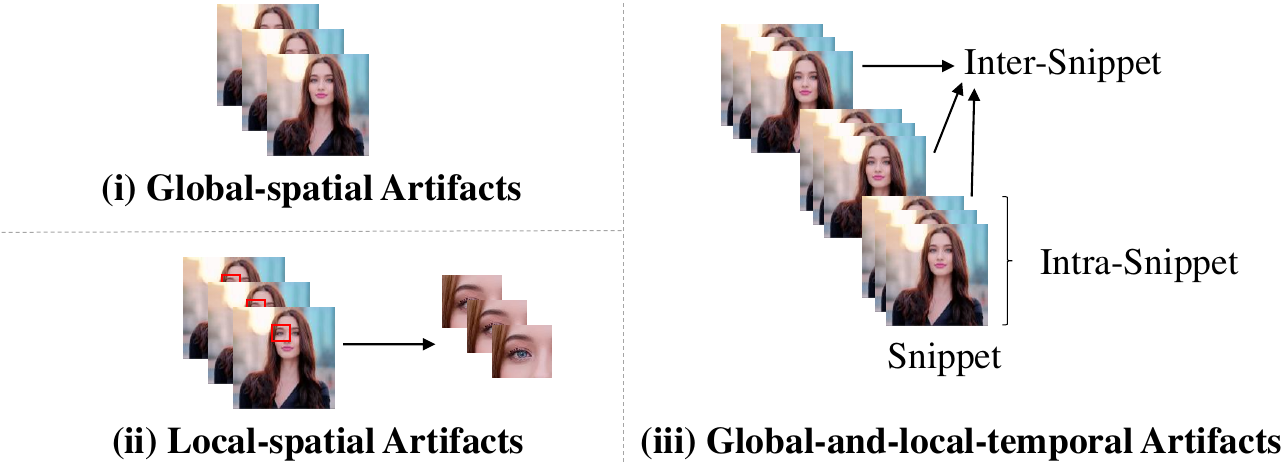}
\end{center}
   \caption{Illustration of global and/or local spatial-temporal artifacts in video-level detection.}
\label{fig:spatial-temporal_artifacts}
\end{figure}

\begin{itemize}
\item Spatial-temporal artifacts. Depending on the size of receptive field across spatial/temporal dimensions, spatial-temporal artifacts can be explored using local or global spatial/temporal receptive fields. As illustrated in Fig.~\ref{fig:spatial-temporal_artifacts}, the commonly explored artifact types include: (i) global-spatial artifacts, (ii) local-spatial artifacts, and (iii) global-and-local-temporal artifacts. In case (iii), a sequence of consecutive frames is grouped into a snippet, and both intra-snippet and inter-snippet artifacts are examined.

\item Audio-visual Artifacts. Generation can introduce an asynchrony between visual and audio signals, causing audio-visual correspondence inconsistencies.

\item Optical Flow-based Artifacts. Optical flow captures the pixel-level motion between adjacent frames~\cite{zhai2021optical} and can be leveraged to detect temporal inconsistencies.

\end{itemize}

\begin{table*}[!t]
\caption{A lookup table for our surveyed papers on detection. These papers are categorized by their detected artifacts and adopted discriminative representation learning strategies. \label{tab:detection_taxonomy}}
\centering
\begin{threeparttable}
      \resizebox{1\textwidth}{!}{
\begin{tabular}{lc|c|c|c|c|c|c}
\hline
\multicolumn{2}{c|}{\textbf{Taxonomy}}&\multicolumn{6}{c}{\textbf{Discriminative Representation Learning Strategies}}\\\cline{1-8}
\multicolumn{2}{c|}{\textbf{\makecell{Generative Artifacts}}} & \multicolumn{1}{c|}{\makecell{Feature Engineering}} & \multicolumn{1}{c|}{\makecell{Multi-task Learning}} & \multicolumn{1}{c|}{\makecell{Transfer Learning}} & \multicolumn{1}{c|}{\makecell{Data$/$Feature Augmentation}} & \multicolumn{1}{c|}{\makecell{Self-supervised Learning}} & \multicolumn{1}{c}{\makecell{Module Design}}\\
\hline
\multirow{10}{*}{\makecell{Image-\\level\\Artifacts}} &
\makecell{Physical\\Characteristics} & \multicolumn{1}{c|}{[\citenum{matern2019exploiting}, \citenum{yang2019exposing}, \citenum{daza2020mebal}, \citenum{zhu2021face}]}&\multicolumn{1}{c|}{[\citenum{zhu2021face}]}&\multicolumn{1}{c|}{--}&\multicolumn{1}{c|}{--}&\multicolumn{1}{c|}{--}&\multicolumn{1}{c}{[\citenum{zhu2021face}]}\\\cline{2-8}
& \makecell{Generation\\Pipeline} & \multicolumn{1}{c|}{\makecell{[\citenum{durall2020watch}, \citenum{liu2021spatial}]}} 
& \multicolumn{1}{c|}{\makecell{[\citenum{li2020face}, \citenum{chen2022self}, \citenum{chen2021local}, \citenum{fei2022learning},\citenum{zhang2022patch},  \citenum{sun2022dual},\\\citenum{zhuang2022uia}, \citenum{dong2022explaining},\citenum{huang2023implicit}, \citenum{dong2022protecting}, \citenum{xia2024mmnet}, \citenum{guo2023hierarchical}]}} 
& \multicolumn{1}{c|}{[\citenum{chen2022ost}]} & \multicolumn{1}{c|}{[\citenum{yang2023progressive}]}
& \multicolumn{1}{c|}{\makecell{[\citenum{li2020face}, \citenum{zhao2021learning}, \citenum{chen2022self}, \citenum{dong2023implicit},\citenum{li2018exposing}, \citenum{shiohara2022detecting},\\\citenum{bai2023aunet}, \citenum{zhang2019detecting},\citenum{jeong2022fingerprintnet}, \citenum{tan2024rethinking}, \citenum{chen2022ost}, \citenum{yu2019attributing}]}} 
&\multicolumn{1}{c}{\makecell{[\citenum{frank2020leveraging}, \citenum{dang2020detection}, \citenum{zhuang2022uia}, \citenum{dong2022explaining},\\\citenum{guo2023hierarchical}, \citenum{huang2022fakelocator}, \citenum{georgiev2023journey}]}}\\\cline{2-8}
& Frequency & \multicolumn{1}{c|}{\makecell{[\citenum{chen2021local}, \citenum{fei2022learning}, \citenum{song2024manifpt}, \citenum{liu2024forgery},\\\citenum{luo2021generalizing}, \citenum{liu2022detecting}, \citenum{gu2022exploiting}]}} 
& \multicolumn{1}{c|}{[\citenum{li2021frequency}]}
&\multicolumn{1}{c|}{[\citenum{woo2022add}]}&\multicolumn{1}{c|}{--}&\multicolumn{1}{c|}{[\citenum{zhou2024freqblender}]}&\multicolumn{1}{c}{\makecell{[\citenum{qian2020thinking}, \citenum{liu2024forgery}, \citenum{wang2022m2tr}, \citenum{gu2022exploiting},\\\citenum{luo2021generalizing}, \citenum{li2021frequency}, \citenum{jeong2022frepgan}, \citenum{2023dynamic}]}}\\\cline{2-8}
& Semantic & \multicolumn{1}{c|}{[\citenum{ojha2023towards}, \citenum{sha2023fake}, \citenum{liu2024forgery}]} &\multicolumn{1}{c|}{[\citenum{bui2022repmix}]}&\multicolumn{1}{c|}{--}&\multicolumn{1}{c|}{[\citenum{bui2022repmix}]}&\multicolumn{1}{c|}{--}&\multicolumn{1}{c}{[\citenum{liu2024forgery}]}\\\cline{2-8}
& Spatial & \multicolumn{1}{c|}{\makecell{[\citenum{tan2023learning}, \citenum{song2024manifpt}, \citenum{huang2023can}, \citenum{asnani2023reverse},\\\citenum{guillaro2023trufor}, \citenum{wang2023dire}, \citenum{ma2023exposing}, \citenum{ricker2024aeroblade}]}}
&\multicolumn{1}{c|}{\makecell{[\citenum{zhao2021multi}, \citenum{li2021frequency}, \citenum{liang2022exploring},\\\citenum{cao2022end}, \citenum{asnani2023reverse}, \citenum{sun2023contrastive}]}} 
&\multicolumn{1}{c|}{\makecell{[\citenum{sun2021domain}, \citenum{kim2021fretal}, \citenum{nadimpalli2022improving},\\\citenum{guarnera2022exploitation}, \citenum{yang2022deepfake}]}}
&\multicolumn{1}{c|}{[\citenum{wang2020cnn}, \citenum{das2021towards}, \citenum{wang2021representative}, \citenum{girish2021towards}]} 
& \multicolumn{1}{c|}{[\citenum{he2021beyond}, \citenum{cao2022end}]}
&\multicolumn{1}{c}{\makecell{[\citenum{zhao2021multi}, \citenum{shao2022detecting}, \citenum{chai2020makes}, \citenum{liu2020global},  \citenum{wang2022m2tr},\\\citenum{sun2022information}, \citenum{luo2021generalizing}, \citenum{liu2022detecting}, \citenum{he2021forgerynet}, \citenum{sun2023contrastive}]}} \\\cline{1-8}
\multirow{9}{*}{\makecell{Video-\\level\\Artifacts}} &
\makecell{Physical\\Characteristics} & \multicolumn{1}{c|}{[\citenum{haliassos2021lips}, \citenum{sun2021improving}, \citenum{qi2020deeprhythm}]} &\multicolumn{1}{c|}{--}&\multicolumn{1}{c|}{--}&\multicolumn{1}{c|}{--}&\multicolumn{1}{c|}{[\citenum{haliassos2021lips}]}& \multicolumn{1}{c}{[\citenum{qi2020deeprhythm}]}\\\cline{2-8}
& Frequency & \multicolumn{1}{c|}{--} &\multicolumn{1}{c|}{--}&\multicolumn{1}{c|}{[\citenum{masi2020two}, \citenum{song2022adaptive}]}&\multicolumn{1}{c|}{--}&\multicolumn{1}{c|}{--}& \multicolumn{1}{c}{[\citenum{masi2020two}, \citenum{song2022adaptive}]} \\\cline{2-8}
& Audio-visual & \multicolumn{1}{c|}{--} &\multicolumn{1}{c|}{[\citenum{zhou2021joint}, \citenum{haliassos2022leveraging}, \citenum{cai2022you}]}&\multicolumn{1}{c|}{--}&\multicolumn{1}{c|}{--}&\multicolumn{1}{c|}{[\citenum{haliassos2022leveraging}, \citenum{feng2023self}]}& \multicolumn{1}{c}{[\citenum{cai2022you}, \citenum{zhang2023ummaformer}]}\\\cline{2-8} 
& Optical Flow & \multicolumn{1}{c|}{--} &\multicolumn{1}{c|}{[\citenum{bai2024ai}]}&\multicolumn{1}{c|}{--}&\multicolumn{1}{c|}{--}&\multicolumn{1}{c|}{--}& \multicolumn{1}{c}{[\citenum{ji2024distinguish}, \citenum{amerini2019deepfake}]}\\\cline{2-8} 
& Global-spatial & \multicolumn{1}{c|}{--} &\multicolumn{1}{c|}{--}&\multicolumn{1}{c|}{}&\multicolumn{1}{c|}{[\citenum{zhang2021detecting}]}&\multicolumn{1}{c|}{[\citenum{hu2022finfer}, \citenum{he2021forgerynet}]}& \multicolumn{1}{c}{[\citenum{song2022adaptive}, \citenum{gu2021spatiotemporal}, \citenum{song2022face}, \citenum{wang2023altfreezing}]}\\\cline{2-8}
& Local-spatial & \multicolumn{1}{c|}{--} &\multicolumn{1}{c|}{--}&\multicolumn{1}{c|}{--}&\multicolumn{1}{c|}{[\citenum{zhang2022deepfake}]}&\multicolumn{1}{c}{--}& \multicolumn{1}{|c}{[\citenum{ijcai2021p102}, \citenum{zheng2021exploring}, \citenum{guan2022delving}, \citenum{chen2024demamba}]} \\\cline{2-8}
& \makecell{Global-and-\\local-temporal} & \multicolumn{1}{c|}{--}&\multicolumn{1}{c|}{[\citenum{gu2022hierarchical}]}&\multicolumn{1}{c|}{--}&\multicolumn{1}{c|}{--}&\multicolumn{1}{c|}{--}& \multicolumn{1}{c}{[\citenum{gu2022delving}, \citenum{gu2022hierarchical}]} \\\cline{2-8}
\hline
\end{tabular}
}
\end{threeparttable}
\end{table*}

\subsubsection{Discriminative Representation Learning Strategies}
Existing detection methods leverage various strategies to capture discriminative representations. This section summarizes the most commonly used strategies in detection.

\begin{itemize}
\item Feature Engineering. Feature engineering~\cite{khurana2018feature} can serve as a preprocessing step in detection, transforming visual media into more distinguishable inputs for learning discriminative representations. These inputs include physical characteristics~\cite{yang2019exposing, zhu2021face}, frequency features~\cite{durall2020watch, gu2022exploiting}, and fixed deep representations~\cite{sha2023fake, ojha2023towards}.

\item Multi-task Learning. Multi-task learning tackles multiple tasks simultaneously by leveraging shared representations, with detection as the main task and other related tasks as auxiliary tasks~\cite{zhang2021survey}. These auxiliary tasks assist in exploring specific artifacts~\cite{chen2021local, fei2022learning} and regularizing feature representations~\cite{zhao2021multi, cao2022end, li2021frequency}.

\item Transfer Learning. Transfer learning transfers knowledge from a source task to improve performance on a target task~\cite{zhang2021survey}, where detection typically serves as the target task. This learning paradigm facilitates learning generalizable features that enable fast adaptation to unseen data distributions. Common incorporated transfer learning strategies include knowledge distillation~\cite{kim2021fretal, woo2022add}, meta-learning~\cite{sun2021domain}, and test-time training~\cite{chen2022ost}.

\item Data$/$Feature Augmentation. Data$/$Feature augmentation enhances the diversity and quantity of training data$/$features, thereby improving model generalizability~\cite{xu2023comprehensive}. Detection methods often incorporate these strategies during data pre-processing or feature extraction. Examples include mixup~\cite{bui2022repmix}, and dropout~\cite{zhang2022deepfake}.

\item Self-supervised Learning. Self-supervised learning trains models using pseudo-labels generated from the training data itself~\cite{jing2020self}. In detection, this strategy assists in generating pseudo-fake samples~\cite{li2020face, zhao2021learning, chen2022self}, guiding pre-task learning~\cite{feng2023self, haliassos2022leveraging}, or guiding auxiliary task learning~\cite{chen2022self}. 

\item Module Design. Detection methods can improve representations through module design, such as incorporating attention module~\cite{dang2020detection, zhuang2022uia} or developing task-specific network architectures~\cite{chai2020makes, liu2020global, liu2022detecting}.

\end{itemize}

\subsection{Forgery Detection}\label{sec:forgery_detection}

Forgery detection performs media-level identification of AI-generated visual media. As illustrated in Fig.~\ref{fig:detection_overview} (i), given a media $M$, forgery detection performs $H_D: M \rightarrow \{0, 1\}$, where $0$ and $1$ typically denotes real and fake media. In this section, following the proposed detection taxonomy, we first review forgery detection methods primarily based on the detected generative artifacts. We then elaborate on these methods based on the incorporated representation learning strategies. A summary of the reviewed methods is presented in Table~\ref{tab:detection_taxonomy}.

\subsubsection{Image-level Artifact-based Detection}
Image-level artifact-based detection detects image-level artifacts to identify forged content.


\paragraph{Physical Characteristics-based Artifacts}
Anomalies in physical characteristics are usually explored in early deepfake detection methods~\cite{matern2019exploiting, yang2019exposing, daza2020mebal}, due to the explicit visual artifacts exhibited in deepfake content. These artifacts include inconsistent eye color, missing reflections, lack of details in the eye and teeth areas~\cite{matern2019exploiting}, inconsistent HeadPose~\cite{yang2019exposing}, 3D decomposed facial details~\cite{zhu2021face}, and unusual eye blinking patterns~\cite{daza2020mebal}. 

\paragraph{Generation Pipeline-based Artifacts}~\label{sec:generation_procedure}
The most commonly explored generation pipelines include \textit{blending operation} and \textit{up-sampling operation}.


The \textit{blending operation} in face swap deepfake generation involves blending regions from different image origins, i.e., the face regions from manipulated images and the background regions from target images. This introduces two types of forgery traces: blending boundary and region inconsistency.

\textit{Blending boundary-related detection} usually serves as an auxiliary task in multi-task learning frameworks to support forgery detection~\cite{li2020face, chen2022self}. These auxiliary tasks include blending boundary segmentation~\cite{li2020face} and prediction of blending type and ratio~\cite{chen2022self}. 

\textit{Region inconsistency} between the face and background regions can serve two roles. First, they can act as attention maps~\cite{dang2020detection}. 
Second, they can supervise auxiliary tasks within multi-task learning framework to guide the learning of specific artifacts. For example, Chen et al.~\cite{chen2021local}, SOLA~\cite{fei2022learning}, Patch Diffusion~\cite{zhang2022patch}, DCL~\cite{sun2022dual}, and UIA-ViT~\cite{zhuang2022uia} use region inconsistency masks to learn local patch inconsistencies. Chen et al.~\cite{chen2022self} and Chen et al.~\cite{chen2021local} employ region inconsistency masks to supervise manipulated region segmentation. FST-Matching~\cite{dong2022explaining}, ICT~\cite{dong2022protecting}, and Huang et al.~\cite{huang2023implicit} utilize such masks to learn identity inconsistency.

Typically, ground-truth masks of region inconsistency are obtained by calculating pixel-level differences between manipulated and target face images. However, such annotations may not be available in public deepfake datasets~\cite{li2020celeb,liu2021spatial}. To address this, some methods exploit \textit{self-supervised pseudo fake samples} by blending real images to simulate blending and regional inconsistency artifacts, providing fine-grained regional annotations. Specifically, Face X-ray~\cite{li2020face}, PCL~\cite{zhao2021learning}, Chen et al.~\cite{chen2022self}, OST~\cite{chen2022ost}, and Dong et al.~\cite{dong2023implicit} generate fake samples by blending real faces with similar facial landmarks~\cite{li2020face, zhao2021learning, chen2022self} or structures~\cite{dong2023implicit}. Other approaches, such as FWA~\cite{li2018exposing}, SBI~\cite{shiohara2022detecting}, and AUNet~\cite{bai2023aunet} create fake samples using a single pristine image. 

The \textit{Up-sampling operation}, widely used in multiple generative models (e.g., VAEs, GANs, DMs), often leaves frequency artifacts, including replicated  frequency spectra~\cite{zhang2019detecting, frank2020leveraging}, as well as anomalies in mid and high frequency band~\cite{zhang2019detecting, durall2020watch}. 
Consequently, classifiers trained on frequency spectra rather than raw RGB pixels tend to be more robust and generalizable.
AutoGAN~\cite{zhang2019detecting} and FingerprintNet~\cite{jeong2022fingerprintnet} use self-supervised fingerprint generators to generate pseudo-fake samples mimicking GAN fingerprints. SPSL~\cite{liu2021spatial} leverages the phase spectrum to detect up-sampling artifacts. Unlike approaches that explore global frequency patterns, NPR~\cite{tan2024rethinking} explores local pixel-level up-sampling traces within the spatial domain.

\paragraph{Frequency Artifacts}
Beyond the frequency artifacts introduced by up-sampling operations, some detection methods design specific modules to explore frequency clues. F3Net~\cite{qian2020thinking} leverages frequency-aware decomposed image components and local frequency statistics to mine the forgery patterns. Li et al.~\cite{li2021frequency} develop an adaptive frequency feature generation module along with a single-center loss to cluster intra-class variations of real images. Luo et al.~\cite{luo2021generalizing} devise a multi-scale high-frequency feature extraction module to complete spatial forgery features. M2TR~\cite{wang2022m2tr} incorporates a frequency filter to explore frequency artifacts. PEL~\cite{gu2022exploiting} learns fine-grained clues from the spatial and frequency domains, followed by self-enhancement and mutual-enhancement modules to improve features from separate domains. FrePGAN~\cite{jeong2022frepgan} adopts a generator to apply frequency-level perturbation maps, making generated images indistinguishable from real images in the frequency domain and thereby mitigating overfitting. Wang et al.~\cite{2023dynamic} develop a two-stream module to explore content-guided features in both spatial and frequency domains, and employ dynamic graph learning to exploit relation-aware features in both domains. ADD~\cite{woo2022add} employs knowledge distillation to perform frequency and multi-view attention distillation, transferring high-frequency information from high-quality to low-quality images. FreqBlender~\cite{zhou2024freqblender} argues that artifacts exist in low-to-high frequencies and accordingly generates pseudo-fake faces by blending different frequency knowledge.

\paragraph{Semantic Artifacts}
Detection methods can utilize large pre-trained vision-language models~\cite{radford2021learning} to identify semantic artifacts. UnivFD~\cite{ojha2023towards} adopts the CLIP image encoder, while DE-FAKE~\cite{sha2023fake} and FatFormer~\cite{liu2024forgery} utilize the CLIP image and text encoder to detect fake images.

\paragraph{Spatial Artifacts}
Spatial artifacts are usually explored in conjunction with different representation learning strategies.

Several methods adopt feature engineering to extract spatial artifacts. LGrad~\cite{tan2023learning}, DIRE~\cite{wang2023dire},  SeDID~\cite{ma2023exposing}, and AEROBLADE~\cite{ricker2024aeroblade} employ pre-trained models to transform images into spatial representations that emphasize discriminative pixels.

Multi-task learning frameworks often incorporate auxiliary tasks to aid in spatial artifact detection. Multiple-attention~\cite{zhao2021multi} designs a regional independence loss and three key module components to exploit fine-grained spatial artifacts. Liang et al.~\cite{liang2022exploring} design reconstruction and content consistency constraint losses to disentangle content information, guiding the detector to focus on spatial artifacts. 

Some detection methods incorporate data$/$feature augmentation.
CNNDetection~\cite{wang2020cnn} applies Gaussian blur and JPEG compression to mitigate overfitting to domain-specific high-frequency cues. Face-Cutout~\cite{das2021towards} and RFM~\cite{wang2021representative} occlude facial regions to prompt the detector to focus on overlooked forgery clues. Nadimpalli et al.~\cite{nadimpalli2022improving} combine supervised and reinforcement learning to train a test-time augmentation policy, mitigating domain shift in cross-dataset scenarios.

He et al.~\cite{he2021beyond} and REECE~\cite{cao2022end} apply self-supervised reconstruction learning, reconstructing real images to enhance the separation between authentic and fake content.

Several methods introduce dedicated modules for spatial artifact detection. Patch-forensics~\cite{chai2020makes} develops patch-based classifiers, followed by an aggregation of patch-wise predictions. Liu et al.~\cite{liu2020global} develop a network that leverages global image texture representations for robust detection. SIA~\cite{sun2022information} designs a plug-and-play self-information attention module to locate informative regions and reaggregate channel-wise feature responses. M2TR~\cite{wang2022m2tr} utilizes a multi-scale self-attention module to capture subtle manipulation artifacts at varying scales. Luo et al.~\cite{luo2021generalizing} devise a residual-guided spatial attention module to highlight spatial manipulation traces. Liu et al.\cite{liu2022detecting} extracts noise patterns and transforms them into the amplitude and phase frequency spectrum, enhancing distinguishable representations. 

Finally, transfer learning approaches are also employed. LTW~\cite{sun2021domain} adopts a meta-learning framework to assign domain-specific weights to improve generalization across multiple domains. FReTAL~\cite{kim2021fretal} employs knowledge distillation to distill feature and prediction distributions, mitigating catastrophic forgetting across varying data distributions.

\subsubsection{Video-level Artifact-based Detection}
Video-level artifact-based detection detects spatial-temporal discontinuity artifacts and multimodality artifacts to perform detection.


\paragraph{Global-spatial Artifacts}
STIL~\cite{gu2021spatiotemporal} explores temporal consistencies along the horizontal and vertical directions of spatial feature maps. TD-3DCNN~\cite{zhang2021detecting} devises a temporal dropout augmentation combined with a 3D Convolutional Neural Network to improve the model's generalization ability. Song et al.\cite{song2022face} design a symmetric transformer to enforce temporal consistency across channel and spatial features, addressing instability in face forgery detection. AltFreezing~\cite{wang2023altfreezing} divides the spatiotemporal model weights into temporal- and spatial-related components, freezing them alternatively during training to prevent the overestimation of one type of artifacts. FInfer~\cite{hu2022finfer} employs an autoregressive model with the representation prediction loss to predict future facial representations based on current frames, distinguishing fake videos. 

\paragraph{Local-spatial Artifacts}
DIANet~\cite{ijcai2021p102} explores global spatial inconsistencies based on structure similarity maps and local spatial inconsistencies focusing on critical regions among adjacent frames. FTCN~\cite{zheng2021exploring} devises a 3D convolution network with a limited spatial convolution kernel size to explore local-spatial temporal artifacts. Zhang et al.~\cite{zhang2022deepfake} perform patch-level spatiotemporal dropout augmentation to diversify the spatiotemporal inconsistency patterns. LTTD~\cite{guan2022delving} models temporal consistency across sequences of restricted spatial regions, and further considers spatial cross-patch inconsistencies to emphasize spatial forgery parts. DeMamba~\cite{chen2024demamba} designs a plug-and-play module based on Mamba~\cite{gu2023mamba} to capture both local and global spatial-temporal inconsistencies.

\paragraph{Global-and-local-temporal Artifacts}
Gu et al.~\cite{gu2022delving} define temporal snippets and explore intra-snippet and inter-snippet inconsistencies. HCIL~\cite{gu2022hierarchical} proposes a hierarchical contrastive learning framework to capture local intra-snippet and global inter-snippet inconsistencies.


\paragraph{Physical Characteristics-based Artifacts}
Anomalies in physical characteristics among sequential frames can indicate fake videos. LipForensics~\cite{haliassos2021lips} learns high-level semantic irregular representations in mouth movements. LRNet~\cite{sun2021improving} leverages precise geometric features to capture temporal artifacts. DeepRhythm~\cite{qi2020deeprhythm} explores sequential features represented by heartbeat rhythm signals along with spatiotemporal features. 

\paragraph{Frequency Artifacts}
Frequency artifacts are usually explored as a complementary perspective to the spatial-temporal artifacts. Masi et al.~\cite{masi2020two} propose a two-branch network to capture temporal artifacts in the spatial and multi-band frequency domain. Song et al.~\cite{song2022adaptive} mine consistency representations across multiple frames from both spatial and frequency domains, followed by an adjustment of false predictions. 

\paragraph{Audio-visual Artifacts}
Using asynchrony between visual and audio signals, Zhou et al.~\cite{zhou2021joint} directly fuse visual-audio pairs jointly for deepfake detection. RealForensics~\cite{haliassos2022leveraging} and Feng et al.~\cite{feng2023self} first learn the temporal synchronization between visual and audio signals with only real data, then perform detection based on the learned cross-modality representations.

\paragraph{Optical Flow-based Artifacts}
Optical flow-based artifacts indicate temporal anomalies and are typically explored jointly with spatial-temporal artifacts~\cite{amerini2019deepfake, bai2024ai, ji2024distinguish}.


\subsection{Fine-grained Forgery Detection}\label{sec:fine-grained_forgery_detection}
In addition to identifying the media-level authenticity, fine-grained forgery detection focuses on detecting manipulations at the pixel, segment, or generator level. This section summarizes mainstream fine-grained forgery detection tasks. An overview of these tasks and their pipelines is illustrated in Fig.~\ref{fig:detection_paper_structure} and Fig.~\ref{fig:detection_overview}. Moreover, we summarize the included methods in Table~\ref{tab:detection_taxonomy}.

\subsubsection{Passive Forgery Localization}\label{sec:passive_forgery_localization}
Passive forgery localization includes both spatial and temporal forgery localization tasks. As illustrated in Fig.~\ref{fig:detection_overview} (ii), given a media $M \in \mathbb{R}^{V}$, where $V$ is the number of pixels in an image or segments in a video, a localization head $H_L$ performs pixel-level or segment-level classification, formulated as $H_L: M \rightarrow \{0, 1\} \in \mathbb{R}^{V}$.

\paragraph{Spatial Forgery Localization}
FakeLocator~\cite{huang2022fakelocator} explores fake textures produced by the upsampling operations and proposes an encoder-decoder network with a face parsing-based attention mechanism to localize these artifacts. HiFi-Net~\cite{guo2023hierarchical} presents a hierarchical fine-grained formulation for image forgery detection and localization representation learning. 
TruFor~\cite{guillaro2023trufor} adopts RGB images and learned noise-sensitive fingerprints to predict pixel-level forgery localization maps.

\paragraph{Temporal Forgery Localization}
ForgeryNet~\cite{he2021forgerynet} adopts existing frame-based~\cite{chollet2017xception} and video-based~\cite{lin2019bmn, lin2018bsn} models for temporal localization. Cai et al.~\cite{cai2022you} propose a 3DCNN model with contrastive, boundary, matching, and frame classification loss functions to explore audio-visual manipulations. UMMAFormer~\cite{zhang2023ummaformer}, designed for video inpainting scenarios, introduces an encoder-decoder architecture for temporal feature reconstruction to enhance temporal differences.

\subsubsection{Forgery Attribution}\label{sec:forgery_attribution}
Forgery attribution aims to identify the generative model or training dataset used to generate media, referred to as model attribution and data attribution, respectively. As illustrated in Fig.~\ref{fig:detection_overview} (iii), the attribution head $H_A$ performs the function $H_A: M \rightarrow C$, where $C = \{c_0, c_1, c_2, ..., c_n\}$ represents a set of $n$ distinct generative models or training datasets.

\paragraph{Model Attribution}
Early approaches focus on \textit{GAN attribution}. Yu et al.~\cite{yu2019attributing} combine image fingerprints with GAN fingerprints for image attribution. Guarnera et al.~\cite{guarnera2022exploitation} leverage prior knowledge of StyleGAN2 model, along with a metric learning framework, to facilitate attribution. DNA-Det~\cite{yang2022deepfake} performs model attribution at the architectural level using patchwise contrastive learning to capture globally consistent fingerprints. RepMix~\cite{bui2022repmix} incorporates semantic feature mixing and mixing ratio prediction to enhance forgery features. Asnani et al.~\cite{asnani2023reverse} reverse-engineer generative model hyperparameters from the corresponding output images. ManiFPT~\cite{song2024manifpt} and Huang et al.~\cite{huang2023can} use transform models to extract manifold-based and generator-specific artifacts. 

Considering most attribution approaches are limited to a closed-set scenario, recent studies have explored \textit{open-set manipulation attribution}\cite{girish2021towards, sun2023contrastive, yang2023progressive}. 
Girish et al.~\cite{girish2021towards} propose an iterative algorithm to discover unseen GAN images from available features. 
POSE~\cite{yang2023progressive} progressively augments closed-set samples to open-set samples to simulate convolution layers in GANs. 
Moreover, \textit{multiple face forgery attribution}~\cite{sun2023contrastive} has been explored by training an attribution model on labeled and unlabeled samples and generalizing to known and unknown manipulated faces, simulating open-world scenes with multiple face forgery methods and types.

\paragraph{Data Attribution}
Recent studies have shown that DMs are prone to memorizing training images and reconstructing them~\cite{schuhmann2022laion, somepalli2023diffusion}. Consequently, Georgiev et al.~\cite{georgiev2023journey} propose attributing DM-generated images back to their training data to detect data leakage and protect copyright. 
D-TRAK~\cite{zheng2023intriguing} evaluates several data attribution approaches and reports that incorporating theoretically unjustified design choices for attribution leads to superior performance. AbC~\cite{wang2023evaluating} evaluates text-to-image model attribution through customization methods. 

\subsubsection{Sequential Manipulation Prediction}\label{sec:sequential_manipulation_prediction}
Sequential manipulation prediction focuses on multi-step manipulated data by predicting a sequential vector of facial manipulation operations and performing a reversal recovery~\cite{shao2022detecting, xia2024mmnet}. Specifically, as illustrated in Fig~\ref{fig:detection_overview} (iv), given media $M$, the prediction head $H_S$ predicts the manipulation sequence, denoted as $H_S: M \rightarrow S$, where $S = \{S_1, S_2, ..., S_s\}$ represents a sequence of $s$ manipulating steps. A recovery decoder $R_S$ then uses both $M$ and the predicted manipulated sequence $S$ to recover an image that closely resembles the pristine image $I$.

Seq-DeepFake~\cite{shao2022detecting} casts the sequential manipulation prediction as an image captioning task employing an autoregressive encoder-decoder model. MMNet~\cite{xia2024mmnet} achieves detection and recovery simultaneously through a multi-stream framework. 

\subsection{Trustworthiness of Detection}\label{sec:trustworthiness_of_detection}
Detection performance may become untrustworthy when encountering specific inputs, such as perturbed inputs~\cite{carlini2020evading, liang2024poisoned} or inputs associated with certain demographic groups~\cite{trinh2021examination, pu2022fairness}, raising trustworthiness concerns. This section reviews key trustworthiness issues of detection, focusing primarily on robustness and fairness.

\subsubsection{Detection Robustness}\label{sec:detection_robustness}
Detection robustness can be threatened by adversarial attacks and backdoor attacks. As shown in Fig.~\ref{fig:detection_overview} (v), adding adversarial perturbations or trigger patterns $\delta$ to forged media can bypass detection, resulting in an erroneous prediction, i.e., $H_D : (M_{fake} + \delta) \rightarrow \{0\}$.

\paragraph{Adversarial Attack}
Adversarial attacks aim to bypass detectors in the inference phase by introducing imperceptible perturbations to input samples.
Based on the level of knowledge about the target detection model, adversarial evasion can be categorized into two types: \textit{detector-based attacks} and \textit{detector-free attacks}. Beyond attacks, several defense strategies have been proposed specifically for robust forgery detection~\cite{chen2023jointly, xia2024inspector}.

\textit{Detector-based attacks} encompass both white-box and black-box attacks. In white-box attacks, adversaries have full access to the model architecture and parameters, while black-box attacks depend on access to training data or model outputs.

In this setting, some studies perform \textit{spatial domain detection evasion}. Carlini et al.~\cite{carlini2020evading} reveal that forensic classifiers are vulnerable to both white-box and black-box adversarial attacks. Hou et al.\cite{hou2023evading} minimize statistical discrepancies between real and fake images to generate adversarial samples. Rather than directly adding perturbations in pixel space, Li et al.\cite{li2021exploring} use gradient descent in the latent space of a generative model to locate adversarial points on the face manifold.
Additionally, Shamshad et al.\cite{shamshad2023evading} guide adversarial generation using either a text prompt or a reference image to create attribute-conditioned adversarial face images.

Other approaches introduce perturbations in both the spatial and frequency domains, leading to \textit{dual domain detection evasion}. Jia et al.\cite{jia2022exploring} use meta-learning to inject perturbations into both domains. Zhou et al.\cite{zhu2023frequency} propose a GAN architecture that fuses dual domain features to produce adversarial examples.
FPBA~\cite{diao2024vulnerabilities} adds adversarial perturbations in various frequency transformation domains and spatial domains from a post-train Bayesian perspective.

\textit{Detector-free attacks} aim to bypass detection systems with minimal knowledge of the detector, requiring neither access to model architecture nor query interactions. These methods typically introduce perturbations to perform \textit{artifact removal}, thereby misleading detectors. FakePolisher~\cite{huang2020fakepolisher} proposes a shallow reconstruction method based on dictionary learning to reduce artifacts. Dong et al.~\cite{dong2022think} correct the power discrepancy to mitigate the frequency spectral artifacts in GAN-generated images, leading to a performance decrease in spectrum-based detectors. Liu et al.~\cite{liu2023making} propose a detector-agnostic trace removal attack that removes spatial anomalies, spectral disparities, and noise fingerprints. 
TraceEvader~\cite{wu2024traceevader} introduces a training-free evasion attack for model attribution in a non-box setting, injecting perturbations into the frequency domain to disrupt trace extraction.

In addition to the aforementioned purpose-built attacks, several studies~\cite{ha2024organic, wang2022deepfake, wu2024watermarks} have revealed that watermarks and perturbations originally intended for copyright protection or generation disruption can also assist in evading detection.

Towards threat posed by adversarial attacks, several defense methods have been developed for robust forgery detection. Chen et al.~\cite{chen2023jointly} design a main and a decoy detector to distinguish the distributional discrepancy of real images. D4~\cite{hooda2024d4} employs multiple sub-networks over disjoint subsets of the frequency spectrum to improve
adversarial robustness. Xia et al.~\cite{xia2024inspector} introduce a coarse-to-fine defense framework to detect multi-grained adversarial perturbations.

\paragraph{Backdoor Attack}
Backdoor attacks in detection aim to embed hidden triggers into models during training, enabling adversaries to mislead detectors at inference by applying specific trigger patterns. Liang et al.~\cite{liang2024poisoned} reveal the backdoor label conflict and trigger pattern stealthiness challenges for backdoor attacks on face forgery detection and propose clean-label backdoor attacks.

\subsubsection{Detection Fairness}\label{sec:detection_fairness}
Detection fairness ensures consistent inference performance across different demographic groups (e.g., gender, age, ethnicity). 

Trinh et al.~\cite{trinh2021examination} and Pu et al.~\cite{pu2022fairness} first reveal detection unfairness across different races and genders of deepfake detectors. Subsequently, several studies~\cite{hazirbas2021towards, nadimpalli2022gbdf} introduce datasets with demographic annotations or balances to support fairness evaluation and promote the development of fairness-aware detectors.



Accordingly, fairness-aware detectors are proposed to ensure consistent performance across demographic groups~\cite{ju2024improving, lin2024preserving}. Ju et al.~\cite{ju2024improving} design novel loss functions to promote fairness in in-domain testing scenarios, where demographic information is either available or unavailable. Lin et al.~\cite{lin2024preserving} focus on fairness in cross-domain evaluations, employing a disentanglement learning module and an optimization module to extract demographic and domain-agnostic forgery features, achieving a flattened loss landscape.



\begin{figure}[t]
\begin{center}
\begin{subfigure}[b]{0.4\textwidth}
    \includegraphics[width=\textwidth]{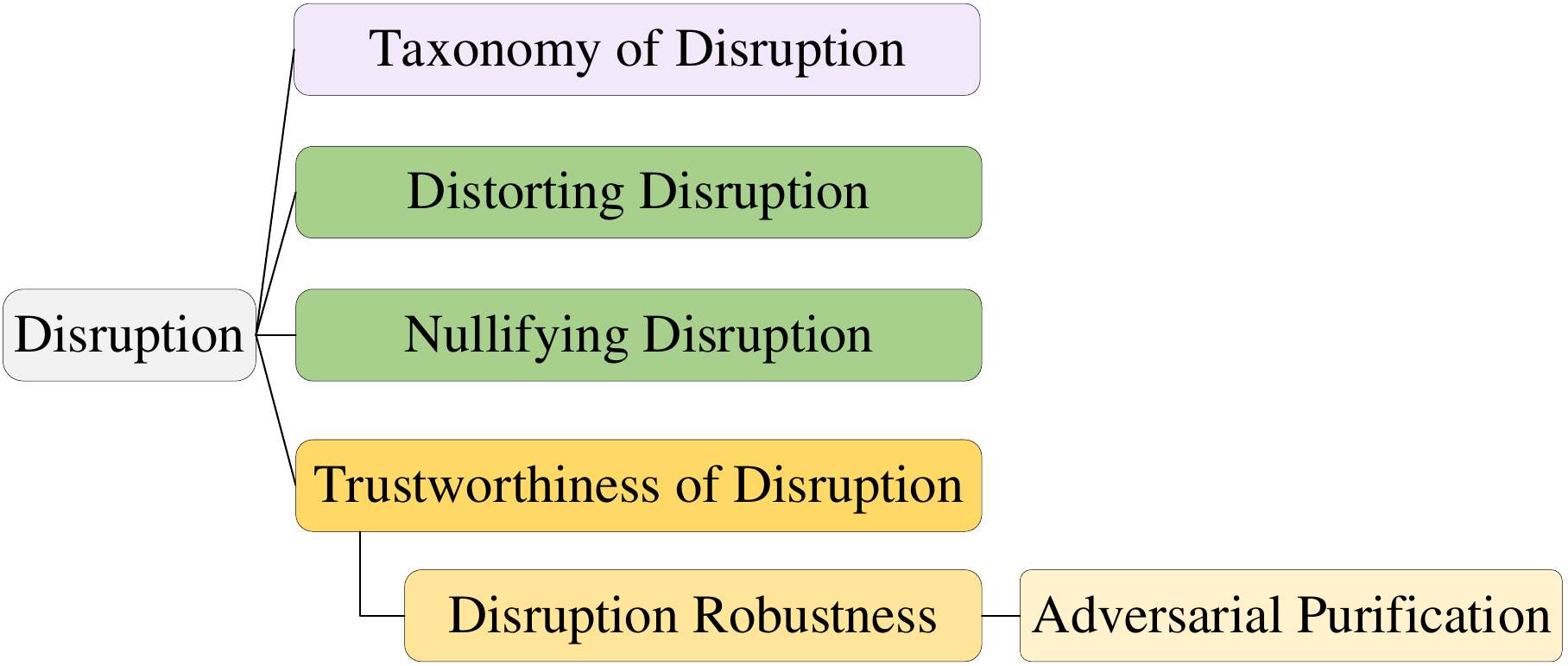}
    \caption{Structure of our paper on disruption.}
    \label{fig:disruption_paper_structure}
\end{subfigure}
\begin{subfigure}[b]{0.5\textwidth}
    \includegraphics[width=\textwidth]{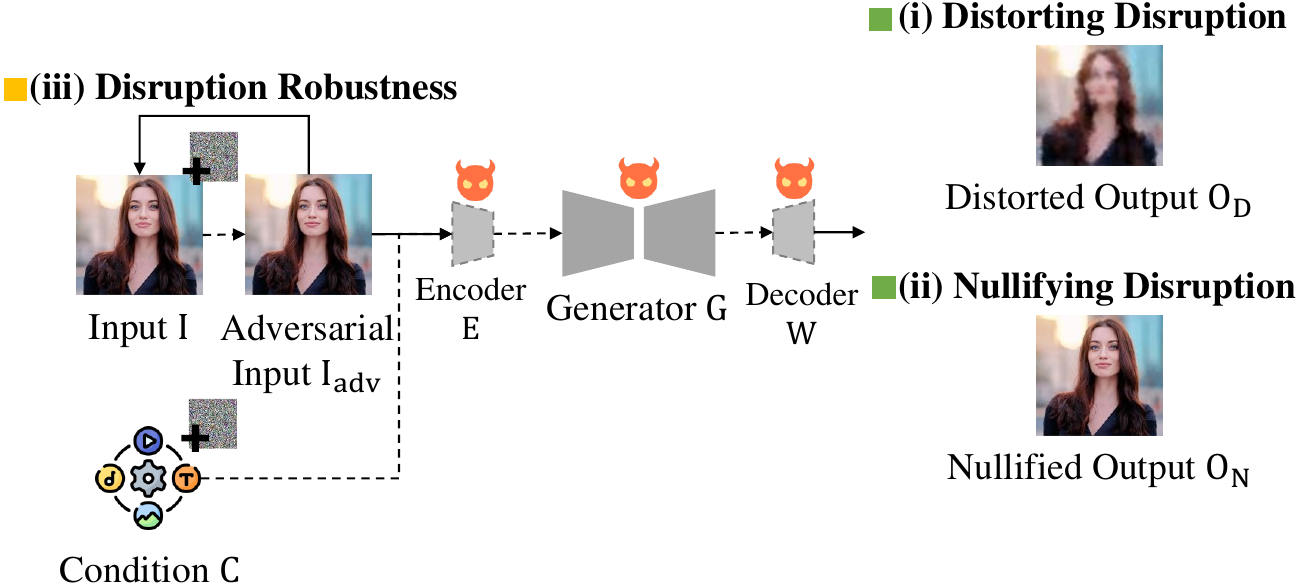}
    \caption{Overview of disruption.}
    \label{fig:disruption_overview}
\end{subfigure}
\end{center}
   \caption{Paper structure and overview of disruption. Disruption can be performed by attacking ($\vcenter{\hbox{\includegraphics[width=8pt]{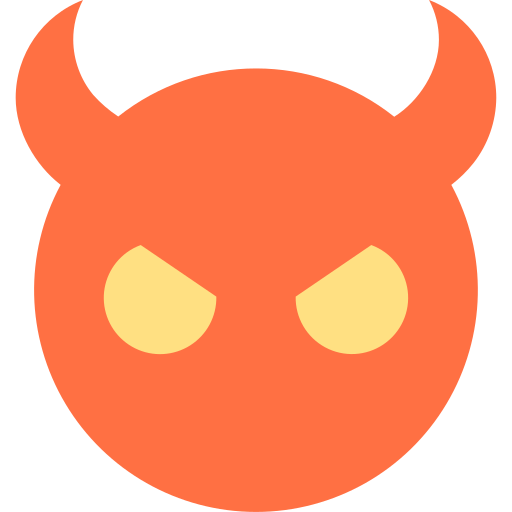}}}$) different modules of generative models, such as the encoder, generator, and decoder, by adding perturbations ($\vcenter{\hbox{\includegraphics[width=8pt]{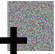}}}$) to the inputs or conditions. Mainstream disruption tasks ($\vcenter{\hbox{\includegraphics[width=8pt]{subtask.png}}}$) include (i) distorting disruption (Section \ref{sec:distorting_disruption}) and (ii) nullifying disruption (Section \ref{sec:nullifying_disruption}). The most studied disruption trustworthiness issue ($\vcenter{\hbox{\includegraphics[width=8pt]{trustworthiness.png}}}$) is (iii) disruption robustness (Section \ref{sec:disruption_robustness}).}
\end{figure}

\section{Disruption}\label{sec:disruption}
Disruption includes distorting and nullifying disruption. As illustrated in Fig.~\ref{fig:disruption_paper_structure}, this section first presents a method taxonomy, followed by a review of representative approaches for each task based on the taxonomy. Finally, a discussion on the trustworthiness of disruption is concluded. An overview of the disruption tasks and their associated trustworthiness is provided in Fig.~\ref{fig:disruption_overview}. 

\subsection{Taxonomy of Disruption}\label{sec:taxonomy_of_disruption}
Disruption is achieved by adding perturbations at different positions in the generation pipeline to attack different generation modules. Accordingly, we categorize existing disruption methods from two perspectives: the attacked module and the perturbation position. It is worth noting that this taxonomy is specific to the generation attack scenario and therefore differs from the general attack categorizations presented in prior work~\cite{ambati2023prat}.

\subsubsection{Perturbation Position}
Disruptive perturbations can be added to the inputs or conditions of generative models.

\begin{itemize}
    \item Input. Perturbations can be added in two manners: (i) directly adding perturbations to the inputs to produce a perturbed input, i.e., $I_{adv} = I + \eta$; (ii) mapping the inputs to representation embeddings, perturbing embeddings, and then reconstructing the perturbed inputs from the perturbed embeddings~\cite{he2022defeating}.

    \item Condition. Perturbations can be added to conditions using a similar approach to that of input perturbations.
    
\end{itemize}

\subsubsection{Attacked Module}
Based on our formulated generation pipeline, the commonly attacked modules include the encoder, generator, and decoder.

\begin{itemize}
    \item Encoder/Decoder. In specific generative scenarios, such as conditional generation~\cite{zhou2021pose} or generation in latent space~\cite{rombach2022high}, attacks can be realized by maximizing the distances between clean and perturbed intermediate feature maps encoded by the encoder/decoder. 
    \item Generator. Attacks on the generator aim to maximize the distances between perturbed and clean intermediate or final generative outputs, where feature maps~\cite{zheng2023understanding} and estimated noise in the reverse process of DMs~\cite{liang2023adversarial, yu2024step} can serve as intermediate generative outputs.
\end{itemize}

\begin{table}[!t]
\centering
\caption{A lookup table for our surveyed papers on disruption. These papers are categorized by the perturbed position and attacked modules for disruption. \label{tab:disruption_taxonomy}}
\resizebox{0.8\textwidth}{!}{
\begin{tabular}{c|ccc}
\hline
\textbf{Taxonomy} & \multicolumn{3}{c}{\textbf{Attacking Module}}\\\cline{1-4}
\textbf{Perturb Position} & \multicolumn{1}{c|}{Encoder} & \multicolumn{1}{c|}{Generator} & \multicolumn{1}{c}{Decoder}\\
\hline
Input & \multicolumn{1}{c|}{[\citenum{salman2023raising}, \citenum{zhang2023robustness}, \citenum{xue2023toward}, \citenum{liang2023mist}, \citenum{ye2023duaw}]} 
& \multicolumn{1}{c|}{\makecell{[\citenum{ruiz2020disrupting}, \citenum{huang2021initiative}, \citenum{yeh2021attack}, \citenum{huang2022cmua}, \citenum{salman2023raising}, \citenum{guan2024adversarial}, \citenum{salman2023raising}, \citenum{zhang2023robustness}, \citenum{xue2023toward}, \\\citenum{xu2024perturbing}, 
\citenum{van2023anti}, \citenum{liang2023adversarial}, \citenum{aneja2022tafim}, \citenum{zhao2023unlearnable}, \citenum{liu2024metacloak}, \citenum{wang2023simac}, \citenum{he2022defeating}, \citenum{zheng2023understanding}]}}
& \multicolumn{1}{c}{[\citenum{salman2023raising}, \citenum{zhang2023robustness}, \citenum{ye2023duaw}]}\\\cline{1-4}
Condition & \multicolumn{1}{c|}{--} & \multicolumn{1}{c|}{[\citenum{yu2024step}]} & \multicolumn{1}{c}{--} \\
\hline
\end{tabular}
}
\end{table}

\subsection{Distorting Disruption}\label{sec:distorting_disruption}
As illustrated in Fig.~\ref{fig:disruption_overview} (i), distorting disruption forces generative models to generate perceptually deteriorated outputs $O_D$. A categorization of these disruption methods is summarized in Table~\ref{tab:disruption_taxonomy}.

Studies focusing on \textit{deepfake distortion} typically add perturbations to the inputs to attack deepfake generators. Yeh et al.~\cite{yeh2020disrupting} are the first to propose distorting and nullifying attacks on image translation GANs, evaluating their effectiveness using proposed similarity and distortion scores. Ruiz et al.~\cite{ruiz2020disrupting} propose a spread-spectrum disruption on GANs. Huang et al.~\cite{huang2021initiative} propose a two-stage training framework with an alternating training strategy between the surrogate manipulation model and perturbation generator, preventing the perturbation generator from trapping in the local optimum. CMUA-Watermark~\cite{huang2022cmua} proposes a cross-model universal attack, including an automatic step size tuning algorithm to find suitable step sizes. TAFIM~\cite{aneja2022tafim} develops a manipulation-aware protection method that guides the manipulation model to produce predefined manipulation targets, with perturbations robust to common image compression. ARS~\cite{guan2024adversarial} designs a two-way protection stream that generates robust perturbations through a single forward process, leveraging perturbation replication and information preservation. 

Recent works have also explored \textit{DMs distortion}, with a particular focus on latent diffusion models (LDMs)\cite{rombach2022high}. These works typically introduce perturbations to the inputs or conditions~\cite{yu2024step} to attack either fixed or fine-tuned diffusion models. Disruption of fixed LDMs can target various components, including the image or condition encoders~\cite{salman2023raising, xue2023toward, zhang2023robustness, liang2023mist}, the generator~\cite{liang2023adversarial, yu2024step, zheng2023understanding}, or the decoder~\cite{zhang2023robustness}. Fine-tuned diffusion models, commonly used for personalized text-to-image generation (e.g., DreamBooth~\cite{ruiz2023dreambooth}), can similarly be disrupted by attacking the encoder~\cite{zhu2024watermark}, the generator (e.g., the denoiser~\cite{zhao2023unlearnable, van2023anti, liu2024metacloak, wang2023simac}, the condition-guided cross-attention module~\cite{xu2024perturbing}), or the decoder~\cite{ye2023duaw}. A slightly different approach is proposed by Zhu et al.~\cite{zhu2024watermark}, who disrupt diffusion models by forcing them to generate images embedded with visible copyright watermarks. Among these methods, several studies~\cite{xue2023toward, ye2023duaw} have found that the encoder is generally more vulnerable than the diffusion model itself. As a result, attacking the encoder proves to be more effective and efficient, as it avoids the multi-step denoising process.

\begin{figure}[t]
\begin{center}
\begin{subfigure}[b]{0.4\textwidth}
    \includegraphics[width=\textwidth]{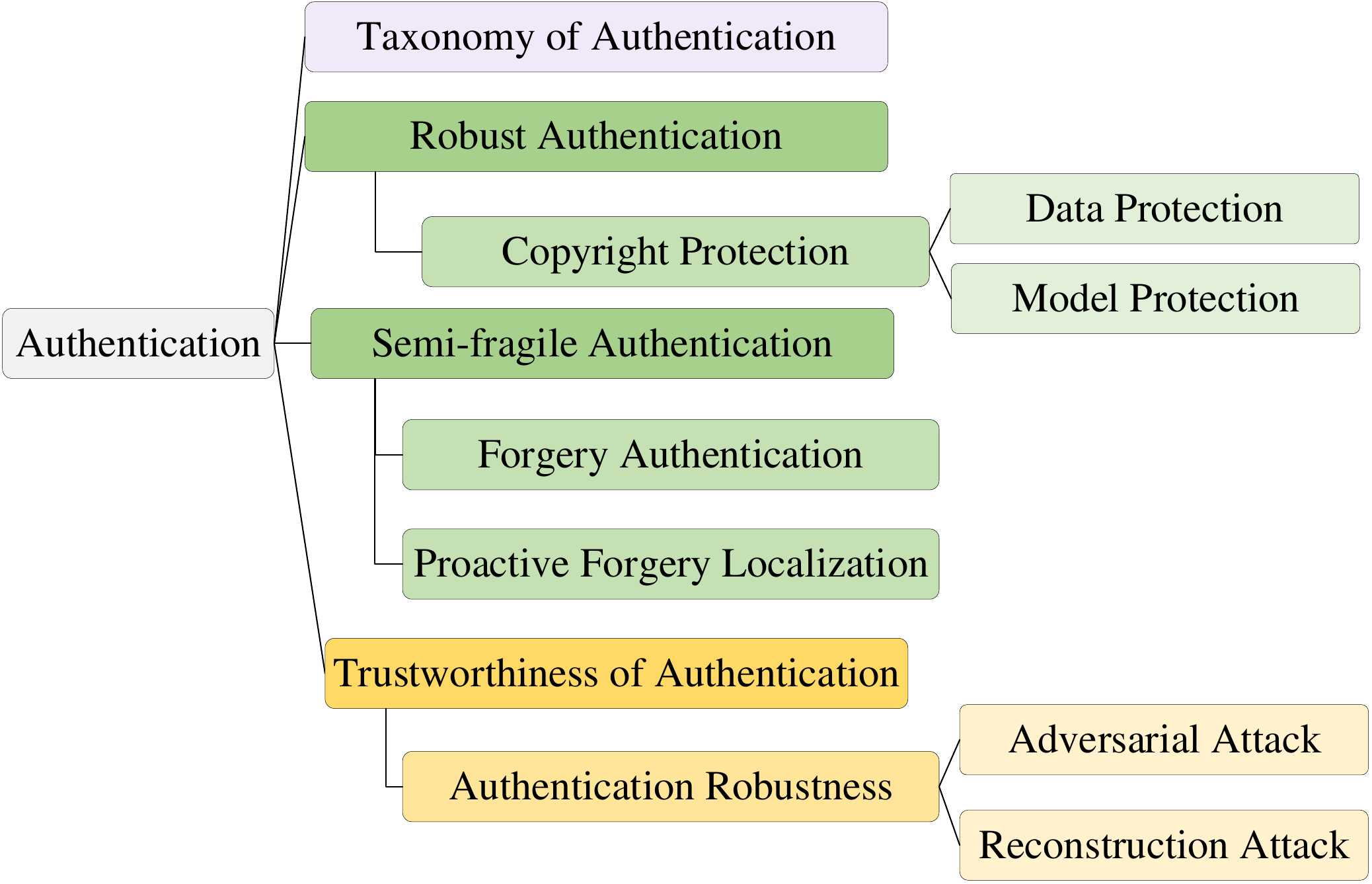}
    \caption{Structure of our paper on authentication.}
    \label{fig:authentication_paper_structure}
\end{subfigure}
\begin{subfigure}[b]{0.5\textwidth}
    \includegraphics[width=\textwidth]{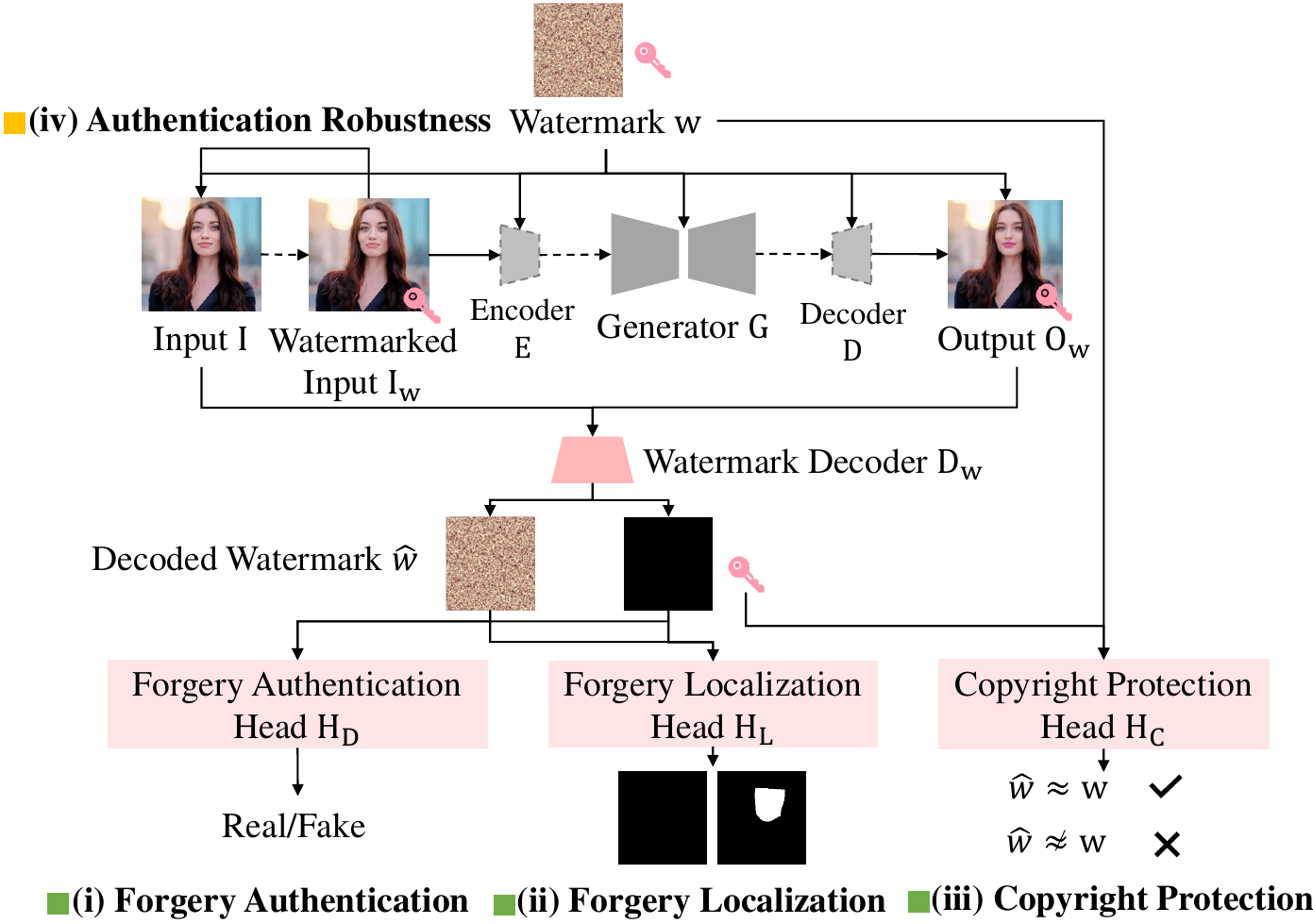}
    \caption{Overview of authentication.}
    \label{fig:authentication_overview}
\end{subfigure}
\caption{Paper structure and overview of authentication. Authentication involves embedding watermarks (robust watermarks for $\vcenter{\hbox{\includegraphics[width=8pt]{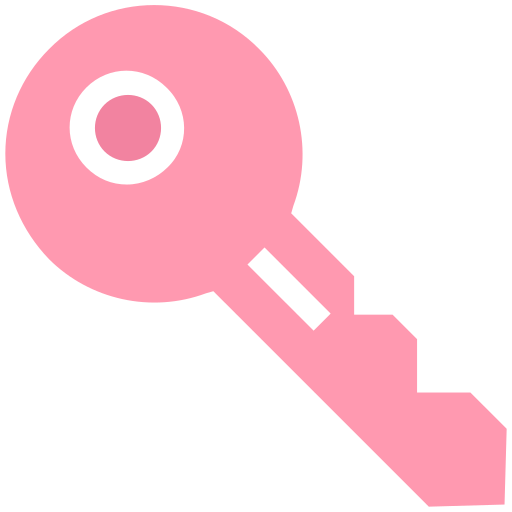}}}$ or semi-fragile watermarks for $\vcenter{\hbox{\includegraphics[width=8pt]{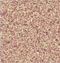}}}$) in synthetic media, followed by watermark decoding and detection for various purposes. These watermarks can be embedded in inputs, encoders, generators, decoders, and generative media. The mainstream authentication tasks ($\vcenter{\hbox{\includegraphics[width=8pt]{subtask.png}}}$) include (i) forgery authentication (Section \ref{sec:forgery_authentication}), (ii) forgery localization (Section \ref{sec:proactive_forgery_localization}), and (iii) copyright protection (Section \ref{sec:copyright_protection}). The most studied authentication trustworthiness issue ($\vcenter{\hbox{\includegraphics[width=8pt]{trustworthiness.png}}}$) is (vi) authentication robustness (Section \ref{sec:authentication_robustness}).}
\end{center}
\end{figure}

\subsection{Nullifying Disruption}\label{sec:nullifying_disruption}
Nullifying disruption aims to guide generative models to generate outputs $O_N$ that are nearly identical to the original inputs, as illustrated in Fig.~\ref{fig:disruption_overview} (ii). A summary of related methods is provided in Table~\ref{tab:disruption_taxonomy}. LaS-GSA~\cite{yeh2021attack} attacks Img2Img GANs using a limit-aware strategy, a gradient-sliding mechanism, and a self-guiding prior. LAE~\cite{he2022defeating} attacks deepfake models by inverting face images into latent codes, searching for adversarial embeddings in their neighbor, and then reconstructing face images from substitute embeddings.

\subsection{Trustworthiness of Disruption}
The most studied trustworthiness issue of disruption is disruption robustness.

\subsubsection{Disruption Robustness}\label{sec:disruption_robustness}
The perturbations added to real images for disruption can be destroyed by adversarial purification. Zhao et al.~\cite{zhao2024can} evaluate disruption effectiveness under complex DMs-fine-tuning scenarios and propose a purification method. Xue et al.~\cite{xue2024pixel} reveal that the vulnerability of LDMs is due to their encoders, while pixel-space diffusion models are robust. To address this, they propose a purifier by applying SDEdit~\cite{meng2021sdedit} in the pixel space. 

\section{Authentication}\label{sec:authentication}
Authentication includes robust and semi-fragile authentication, differentiated by the fragility levels of the embedded watermarks. Watermarks for robust authentication are designed to withstand manipulations by generative models and are primarily applied for copyright protection. In contrast, watermarks for semi-fragile authentication are destroyed after generation, creating discrepancies between real and generated media, and are used for forgery authentication and localization. As shown in Fig.~\ref{fig:authentication_paper_structure}, the method taxonomy is first introduced in this section, followed by an overview of methods for each task. Finally, we review the trustworthiness of the authentication defense. An overview of authentication-related tasks and their trustworthiness is illustrated in Fig.~\ref{fig:authentication_overview}.  

\subsection{Taxonomy of Authentication}
Authentication involves embedding various types of watermarks at different positions within the generation pipeline. Therefore, we categorize existing authentication methods from two perspectives: the watermark position and the watermark type.

\subsubsection{Watermark Position}
Watermarks can be embedded into the inputs, input embeddings, generator, or generative media. Since watermarking methods applied directly to generative media often use off-the-shelf techniques~\cite{ai_responsibly}, we do not elaborate on them here.

\begin{itemize}
    \item Input. Watermarks can be embedded by: (i) directly adding them to the inputs; (ii) embedding them into input representation embeddings, followed by reconstruction.
    \item Decoder. For generation in latent space~\cite{rombach2022high}, watermarks can be injected into the decoder~\cite{fernandez2023stable} or their intermediate generative products, such as feature maps produced by the decoder~\cite{xiong2023flexible}.
    \item Generator. Watermarks can be embedded into generators or their produced intermediate generative products~\cite{wen2023tree}.
\end{itemize}

\subsubsection{Watermark Type} Tree types of watermarks are commonly employed: (i) Bit string $w \in \{0, 1\}^k$, where $k$ is the bit size; (ii) Spatial-domain watermarks $w$, embedded in the spatial domain, where $w \in \mathbb{R}^{V}$ and $V$ is the number of pixels in an image or feature map; (iii) Frequency-domain watermarks, embedded in the Fourier space.

\begin{table}[!t]
\caption{A lookup table for our surveyed papers on authentication. These papers are categorized by their embedded watermark position and type for authentication.\label{tab:authentication_taxonomy}}
\centering
\begin{threeparttable}
\resizebox{1\textwidth}{!}{
\begin{tabular}{c|ccc}
\hline
\textbf{Taxonomy} & \multicolumn{3}{c}{\textbf{Watermark Type}} \\\cline{1-4}
\textbf{Watermark Position} & \multicolumn{1}{c|}{Bit String} & \multicolumn{1}{c|}{Spatial Watermark} & \multicolumn{1}{c}{Frequency Watermark}\\
\hline
Input & \multicolumn{1}{c|}{[\citenum{yu2021artificial}, \citenum{wang2021faketagger}, \citenum{zeng2023securing}, \citenum{lukas2023ptw}, \citenum{wu2023sepmark}, \citenum{zhao2023proactive}, \citenum{wu2023sepmark}, \citenum{neekhara2022facesigns}, \citenum{zhang2024editguard}]} & \multicolumn{1}{c|}{[\citenum{cui2023diffusionshield}, \citenum{ma2023generative}, \citenum{asnani2022proactive}, \citenum{asnani2023malp}, \citenum{beuve2023waterlo}, \citenum{cui2025ft}]} & \multicolumn{1}{c}{--}\\\cline{1-4}
Generator & \multicolumn{1}{c|}{--} & \multicolumn{1}{c|}{--} & \multicolumn{1}{c}{[\citenum{wen2023tree}]} \\\cline{1-4}
Decoder & \multicolumn{1}{c|}{[\citenum{fernandez2023stable}, \citenum{xiong2023flexible}, \citenum{kim2024wouaf}]} & \multicolumn{1}{c|}{--} & \multicolumn{1}{c}{--} \\\cline{1-4}
\hline
\end{tabular}
}
\end{threeparttable}
\end{table}

\subsection{Robust Authentication}\label{sec:robust_authentication}
Robust authentication involves embedding watermarks that are resistant to the generation process, providing copyright protection by comparing the similarity of decoded watermarks to the original watermarks held by the copyright owner, as shown in Fig.~\ref{fig:authentication_overview} (iii). The corresponding methods are summarized in Table~\ref{tab:authentication_taxonomy}.

\subsubsection{Copyright Protection}\label{sec:copyright_protection}
Copyright protection aims to protect the source training data or the generative models used to produce generative media. Specifically, a copyright owner embeds a robust watermark $w$ before, during, or after generation. The generative model then produces an output $O_w$ that contains the embedded, transferable watermark. A decoded watermark $\hat{w}$ can subsequently be extracted from $O_w$. By comparing $\hat{w}$ with $w$, ownership can be verified if the matching score exceeds a predefined threshold $\tau$.

\textit{Source data protection} typically embeds watermarks into source training data (inputs) of generative models. Early works focus on \textit{GAN watermarking}. Yu et al.~\cite{yu2021artificial} trains an encoder and decoder independent of the GAN training to embed bit string watermarks. FakeTagger~\cite{wang2021faketagger} designs an encoder-decoder watermarking framework that injects bit string watermarks into the channel codings of input representations, followed by input reconstruction. Recent works also concentrate on \textit{DM watermarking}, with some tailored for training-from-scratch settings~\cite {cui2023diffusionshield}, and others for fine-tuning settings~\cite {ma2023generative, cui2025ft}. DiffusionShield~\cite{cui2023diffusionshield} introduces watermarks with pattern uniformity and employs joint optimization to improve the reproduction of watermarks in generated images. GenWatermark~\cite{ma2023generative} adds spatial watermarks into training data and incorporates a detector that is fine-tuned along with the image synthesis process to decode the watermarks. FT-Shield~\cite{cui2025ft} embeds watermarks at the early stage of fine-tuning and constructs a mixture of detectors effective across various fine-tuned models. 

\textit{Model protection} signs the generative model used to produce generative media, where traceable watermarks can be embedded at any of the previously discussed watermarking positions. 
Input watermarking typically requires retraining generative models to produce watermarks via the generator~\cite{zeng2023securing, yu2021artificial, lukas2023ptw}. 
For generator watermarking, Tree-Ring~\cite{wen2023tree} embeds frequency-domain watermarks into the initial noise vectors of DMs and decodes them by inverting the diffusion process. 
Decoder watermarking usually requires partial retraining.
Fernandez et al.~\cite{fernandez2023stable} fine-tune the latent decoder of LDMs under the guidance of a watermark decoder to embed bit string watermarks. Xiong et al.~\cite{xiong2023flexible} encode bit string watermarks into a message matrix and embed it into the intermediate feature maps of the decoder. WOUAF~\cite{kim2024wouaf} encodes watermarks via a mapping network and embeds them into decoders via weight modulation. 

\subsection{Semi-fragile Authentication}
As illustrated in Fig.~\ref{fig:authentication_overview} (i) and (ii), semi-fragile authentication embeds watermarks $w$ in the inputs $I$, while subsequent generation destroys $w$ and generates an output $O_w$ with destroyed watermarks. By comparing the original watermark $w$ with the extracted watermark $\hat{w}$, forgery authentication and localization can be conducted. These methods are summarized in Table~\ref{tab:authentication_taxonomy}.

\subsubsection{Forgery Authentication}\label{sec:forgery_authentication}
Forgery authentication detects forged watermarks $\hat{w}$ with a detection head $H_D$, denote as $H_D: \hat{w} \rightarrow \{0, 1\}$, where $\hat{w}=D_W(G(I, w))$. 

Watermarks for forgery authentication are typically embedded into inputs. Asnani et al.~\cite{asnani2022proactive} estimate a set of spatial watermarks under specific constraints to encrypt real images. SepMark~\cite{wu2023sepmark} proposes a framework with a watermark encoder and two decoders with different levels of decoding robustness for source tracing and detection. Zhao et al.~\cite{zhao2023proactive} embed bit string watermarks into the facial identity features to detect Deepfake manipulation. AdvMark~\cite{wu2024watermarks} proposes adversarial bit string watermarking to enhance image detectability without compromising provenance tracking. FaceSigns~\cite{neekhara2022facesigns} embeds bit string watermarks into real images for deepfake detection.

\subsubsection{Proactive Forgery Localization}\label{sec:proactive_forgery_localization}
Proactive forgery localization uses a localization head $H_L$ to predict pixel-level classification results based on the decoded watermarks $\hat{w}$, denoted as $H_L: \hat{w} \rightarrow \{0, 1\} \in \mathbb{R}^{V}$, where $\hat{w}=D_W(G(I, w)) \in \mathbb{R}^{V}$. 

These methods typically embed spatial watermarks into inputs to perform proactive forgery localization. MaLP~\cite{asnani2023malp} watermarks real images with a learnable template that captures both local and global features, enabling manipulation region and binary classification on both GMs and GANs. EditGuard~\cite{zhang2024editguard} embeds both a robust bit string and a semi-fragile spatial watermark for simultaneous copyright protection and tamper localization using a joint image-bit steganography technique. Waterlo~\cite{beuve2023waterlo} applies a semi-fragile watermark that disappears in modified regions to identify and localize image modifications.

\subsection{Trustworthiness of Authentication}
Research on the authentication trustworthiness primarily focuses on authentication robustness.

\subsubsection{Authentication Robustness}\label{sec:authentication_robustness}
Authentication relies on embedding watermarks in AI-generated content for subsequent verification. However, these watermarks can be compromised by traditional distortions (e.g., JPEG compression and cropping~\cite{fernandez2023stable}) or destroyed by \textit{reconstruction attacks} and \textit{adversarial attacks}. 

For \textit{reconstruction attacks}, Zhao et al.~\cite{zhao2024invisible} and Saberi et al.~\cite{saberi2023robustness} add noise to watermarked images and then reconstruct images by generative models. For \textit{adversarial attacks}, PTW~\cite{lukas2023ptw} reveals that watermarking schemes are vulnerable against adaptive white-box attacks. WEvade~\cite{jiang2023evading} leverages an adversarial post-processing layer to attack watermark detectors under both white-box and black-box settings. Lukas et al.~\cite{lukas2023leveraging} apply adaptive attacks by creating surrogate keys to attack the authentication system. WAVES~\cite{an2024benchmarking} benchmarks the robustness of watermarks against multiple types of attacks. According to their evaluation, different watermarking methods exhibit different vulnerabilities across adversarial scenarios, and methods incorporating augmentation or adversarial training tend to demonstrate greater robustness. Moreover, a trade-off exists between watermark robustness and quality, which should be carefully considered during watermarking design.

\begin{table*}[!t]
\caption{An overview of popular Deepfake datasets is presented, summarizing the publication year, media modalities, total number of media samples, number of subjects, number of manipulation methods per Deepfake type, inclusion of multiple faces, availability of an additional test set, number of perturbation types, presence of multi-modal data, and the number of annotation types provided by each dataset. \label{tab:deepfake_datasets}}
\centering
      \resizebox{1\textwidth}{!}{
\begin{tabular}{l|c|c|cc|c|cccc|c|c|c|c|c}
\hline
\multirow{3}{*}{Dataset} & 
\multirow{3}{*}{Year}  & 
\multirow{3}{*}{\makecell{Media\\Modality}} &
\multicolumn{2} {c|}{Total Media} & 
\multirow{3}{*}{\makecell{Total\\Subj}} & 
\multicolumn{4}{c|}{Deepfake Types} & 
\multirow{3}{*}{\makecell{Multi-\\face}} & 
\multirow{3}{*}{\makecell{Extra\\Test Set}}& 
\multirow{3}{*}{\makecell{Perturb}}&
\multirow{3}{*}{\makecell{Multi-\\Modal}}&
\multirow{3}{*}{Anno}\\
&&&\multirow{2}{*}{Real} & \multirow{2}{*}{Fake} & & \makecell{Entire\\Syn.} & \makecell{Attri.\\Mani.} & \makecell{Id.\\Swap} & \makecell{Face\\Re.} & & & & \\[0.2em]
\hline
UADFV~\cite{li2018ictu} & 2018 & Video & 49 & 49 & 49 & -- & -- & 1 & -- & -- & -- & -- & -- & 1\\[0.2em]
DF-TIMIT~\cite{korshunov2018deepfakes} & 2018 & Video & 640 & 320 & 32 & -- &  -- & 1 & -- & -- & -- & 1 & Audio & 1 \\[0.2em]
FaceForensics++~\cite{rossler2019faceforensics++} & 2019 & Video & 1,000 & 4,000 & 977 & -- & -- & 3 & 2 & -- & -- & 1 & -- & 2 \\[0.2em]
DFD~\cite{DDD_GoogleJigSaw2019} & 2019 & Video & 363 & 3,068 & 28 & -- & -- & 5 & -- & -- & -- & -- & -- & 1\\[0.2em]
DFDC-preview~\cite{dolhansky2019deepfake} & 2019 & Video & 4,073 & 1,140 & 66 & -- & -- & 2 & -- & -- & -- & 3 & -- & 1 \\[0.2em]
Celeb-DF-v2~\cite{li2020celeb} & 2020 & Video & 590 & 5,639 & 59 & -- & -- & 1 & -- & -- & -- & -- & -- & 1\\[0.2em]
WildDeepfake~\cite{zi2020wilddeepfake} & 2020 & Video & 3,805 & 3,509 & -- & unk. &  unk. & unk. & unk. & -- & -- & -- & -- & 1\\[0.2em]
DeeperForensics-1.0~\cite{jiang2020deeperforensics} & 2020 & Video & 50,000 & 10,000 & 100 & -- & -- & -- & 1 & -- & \checkmark & 7 & -- & 1 \\[0.2em]
DFFD~\cite{dang2020detection} & 2020 & Video & 59,703 & 243,336 & -- & 2 & 2 & 3 & 1 & -- & -- & -- & -- &  1 \\[0.2em]
DFDC~\cite{dolhansky2020deepfake} & 2020 & Video & 104,500 & 23,654 & 960 & -- & -- & 8 & -- & -- & \checkmark & 19 & Audio & 1 \\[0.2em]
ForgeryNet~\cite{he2021forgerynet} & 2021 & V + I & 99,630 &  121,617 & 5,400 & -- & 5 & 7 & 3 & -- & \checkmark & 36 & -- & 4\\[0.2em]
FakeAVCeleb~\cite{khalid2021fakeavceleb} & 2021 & Video & 500 & 19,500 & 500 & -- & -- & 2 & 1 & -- & -- & -- & Audio & 2 \\[0.2em]
FFIW~\cite{zhou2021face} & 2021 & Video & 10,000 & 10,000 & -- & -- & -- & -- & 3 & \checkmark & -- & -- & -- & 2 \\[0.2em]
OpenForensics~\cite{le2021openforensics} & 2021 & Image & 45,473 & 70,325 & -- & -- & -- & -- & 2 & \checkmark & \checkmark & 6 & --& 5\\[0.2em]
KoDF~\cite{kwon2021kodf} & 2021 & Video & 62,166 & 175,776 &  403 & -- & -- & 3 & 3 & -- & \checkmark & -- & Audio & 1 \\[0.2em]
LAV-DF~\cite{cai2022you} & 2022 & Video & 36,413 & 99.873 & 153 & -- & -- & -- & 1 & -- & -- & -- & Audio & 2\\[0.2em]
DF-Platter~\cite{narayan2023df} & 2023 & Video & 764 & 132,496 & 454 & -- & -- & 3 & -- & \checkmark & \checkmark & 1 & -- & 6 \\[0.2em]
AV-Deepfake1M~\cite{cai2024av} & 2024 & Video & 286,721 & 860,039 & 2,068 & -- & -- & -- & 1 & -- & -- & -- & Audio & 4\\[0.2em]
\hline
\end{tabular}
}
\begin{threeparttable}
\begin{tablenotes}
      \scriptsize
      \item[*] V + I: The dataset contains videos and images.
      \item[*] "Entrie Syn.", "Attri. Mani.", "Id. Swap", and "Face Re." refer to entire face synthesis, attribute manipulation, identity swap, and face reenactment, respectively. 
      \item[*] "unk." represents unknown.
    \end{tablenotes}
\end{threeparttable}
\end{table*}

\begin{table*}[!t]
\caption{An overview of popular universal generative visual media datasets, which go beyond face manipulations in deepfake datasets. The public availability is additionally summarized here.\label{tab:universal_datasets}}
\centering
      \resizebox{1\textwidth}{!}{

\begin{tabular}{l|c|c|cc|ccc|c|c|c|c|c}
\hline
\multirow{2}{*}{Dataset} & 
\multirow{2}{*}{Year}  & 
\multirow{2}{*}{\makecell{Media \\ Modality}} &
\multicolumn{2}{c|}{Total Media} & 
\multicolumn{3}{c|}{Generator Types} & 
\multirow{2}{*}{\makecell{Public \\ Avail}} & 
\multirow{2}{*}{\makecell{Extra \\ Test Set}}& 
\multirow{2}{*}{\makecell{Perturb}}&
\multirow{2}{*}{\makecell{Multi- \\ Modal}}&
\multirow{2}{*}{Anno}\\
&&&Real & Fake & GAN & Diffusion & Others & & & & & \\[0.2em]
\hline
CNNDetection~\cite{wang2020cnn} & 2020 & Image & 362,000 & 362,000 &  6 & -- & 5 & \checkmark & -- & -- & -- & 2 \\[0.2em]
PAL4Inpaint~\cite{zhang2022perceptual} & 2022 & Image & -- & 4,795 & 3 & -- & -- & \checkmark & -- & -- & -- & 2 \\[0.2em]
HiFi-IFDL~\cite{guo2023hierarchical} & 2023 & Image &100,000 & 100,000 & 4 & 4 & 5 & Part & -- & 2 & -- & 3 \\[0.2em]
DIRE~\cite{wang2023dire} & 2023 & Image &232,000 & 232,000 & -- & 10 & -- & \checkmark & -- & -- & -- & 2\\[0.2em]
DE-FAKE~\cite{sha2023fake} & 2023 & Image &20,000 &  60,000 & -- & 4 & -- & -- & -- & -- & Text & 2\\[0.2em]
AutoSplice~\cite{jia2023autosplice} & 2023 & Image &3,621 & 2,273 & -- & 1 & -- & \checkmark & -- & 1 & Text & 3\\[0.2em]
PAL4VST~\cite{zhang2023perceptual} & 2023 & Image &-- & 10,168 & 6 & 5 & 1 & \checkmark & \checkmark & -- & -- & 2 \\[0.2em]
GenImage~\cite{zhu2024genimage} & 2024 & Image & 1,331,167 & 1,350,000 & 1 & 7 & -- & \checkmark & -- & 3 & -- & 2 \\[0.2em]
GenVideo~\cite{chen2024demamba} & 2024 & Video & 1,223,511 & 1,078,838 & 1 & 9 & 4 & -- & -- & 8 & -- & 1 \\[0.2em]
\hline
\end{tabular}
}
\end{table*}

\begin{table*}[!t]
\caption{Comparison of deepfake detection methods. The publication year, belonging categories according to the taxonomy introduced in Section~\ref{sec:detection}, backbone networks, detection scenarios, and detection performance of each method are summarized in the table. To enable a fair comparison of detection performance, we merely summarize the commonly used detection scenarios and evaluation metrics in the field of deepfake detection here. Metric-level refers to the unit of samples on which the performance is calculated. \label{tab:deepfake_detection_method_comparison}}
\centering
\begin{threeparttable}
      \resizebox{1\textwidth}{!}{
\begin{tabular}{l|c|c|c|c|c|c|c|c|c|c}
\hline
\multirow{3}{*}{Method} & \multirow{3}{*}{Year} & \multicolumn{2}{c|}{\multirow{2}{*}{Taxonomy}} & \multirow{3}{*}{Backbone} & \multirow{3}{*}{\makecell{Training\\Dataset}} & \multirow{3}{*}{\makecell{Metric\\Level}} & \multicolumn{4}{c}{Performance (AUC\%)} \\
& & \multicolumn{2}{c|}{} & & &  & In-domain & \multicolumn{3}{c}{Cross-domain} \\
& & Modal & Subcategory & & &  & FF++ & DFD & CD2 & DFDC\\
\hline
Zhu et al.~\cite{zhu2021face} & 2021 & \multirow{32}{*}{Image}& PC; FE, MTL, MD & XceptionNet~\cite{chollet2017xception} & FF++-HQ & Frame & 99.45 & 78.65 & -- & 66.09 \\
SPSL~\cite{liu2021spatial} & 2021 &  & GP; FE & XceptionNet~\cite{chollet2017xception} & FF++-HQ & Frame & 95.32 & -- & 76.88 & -- \\
Face X-ray~\cite{li2020face} & 2020 &  & GP; MTL, SSL & HRNet~\cite{sun2019deep} & FF++-unk. & Frame & 98.52 & 95.40 & 80.58 & \makecell{80.92\\(DFDC-P)} \\
Chen et al.~\cite{chen2022self} & 2022 & & GP; MTL, SSL & XceptionNet~\cite{chollet2017xception} & FF++-HQ & Frame & 98.4 & -- & 79.7 & -- \\
Chen et al.~\cite{chen2021local} & 2021 &  & GP, Freq; MTL, FE & unk. & FF++-HQ & Frame & 99.46 & 89.24 & 78.26 & 76.53 \\
SOLA~\cite{fei2022learning} & 2022 &  & GP, Freq; MTL, FE & ResNet~\cite{he2016deep} & FF++-DF-HQ & Frame & 100.0 & -- & 76.02 & --\\
Zhang et al.~\cite{zhang2022patch} & 2022 &  & GP; MTL & EfficientNet~\cite{tan2019efficientnet} & FF++-HQ & Frame & 98.51 & 97.38 & 82.25 & -- \\
DCL~\cite{sun2022dual} & 2022 &  & GP; MTL & EfficientNet~\cite{tan2019efficientnet} & FF++-HQ & Frame & 99.30 & 91.66 & 82.30 & 76.71 \\
UIA-ViT~\cite{zhuang2022uia} & 2022 &  & GP; MTL, MD & ViT~\cite{dosovitskiy2020image} & FF++-HQ & Frame, Video & 99.33 & 94.68 & 82.41 & 75.80 \\
Dong et al.~\cite{dong2022explaining} & 2022 &  & GP; MTL, MD & EfficientNet~\cite{tan2019efficientnet} & FF++-HQ & Video & 98.81 & -- & 89.39 & -- \\
Huang et al.~\cite{huang2023implicit} & 2023 &  & GP; MTL & ResNet~\cite{he2016deep} & FF++-HQ & Frame & 99.32 & 93.92 & 83.80 & 81.23 \\
ICT-Ref~\cite{dong2022protecting} & 2022 &  & GP; MTL & ViT~\cite{dosovitskiy2020image} & FF++-unk. & Frame & 98.56 & 93.17 & 94.43 & -- \\
OST~\cite{chen2022ost} & 2022 &  & GP; TL, SSL & XceptionNet~\cite{chollet2017xception} & FF++-HQ & Frame & 98.2 & -- & 74.8 & 83.3\\
PCL~\cite{zhao2021learning} & 2021 &  & GP; SSL & ResNet~\cite{he2016deep} & FF++-raw & Video & 99.79 & 99.07 & 90.03 & 67.52 \\
Dong et al.~\cite{dong2023implicit} & 2023 &  & GP; SSL & EfficientNet~\cite{tan2019efficientnet} & FF++-unk. & Video & 99.79 & -- & 93.88 & 73.85 \\
SBI~\cite{shiohara2022detecting} & 2022 &  & GP; SSL & EfficientNet~\cite{tan2019efficientnet} & FF++-raw & Video & 99.64 & 97.56 & 93.18 & 72.42 \\
AUNet~\cite{bai2023aunet} & 2023 &  & GP; SSL & XceptionNet~\cite{chollet2017xception} & FF++-HQ & Frame & 99.89 & 99.22 & 92.77 & 73.82 \\
Luo et al.~\cite{luo2021generalizing} & 2021 &  & Freq, Spatial ; FE, MD & XceptionNet~\cite{chollet2017xception} & FF++-HQ & Frame & -- & 91.9 & 79.4 & 79.7 \\
Gu et al.~\cite{gu2022exploiting} & 2022 &  & Freq; FE, MD & EfficientNet~\cite{tan2019efficientnet} & FF++-LQ & Frame & 94.28 & 75.86 & 69.18 & 63.31 \\
Li et al.~\cite{li2021frequency} & 2021 &  & Freq, Spatial; MTL, MD & XceptionNet~\cite{chollet2017xception} & FF++-LQ & Video & 91.6 & --& --& --\\
F3-Net~\cite{qian2020thinking} & 2020 &  & Freq; MD & XceptionNet~\cite{chollet2017xception} & FF++-HQ & Video & 99.3 & -- & -- & --\\
M2TR~\cite{wang2022m2tr} & 2022 &  & Freq, Spatial; MD & EfficientNet~\cite{tan2019efficientnet} & FF++-HQ & Frame & 99.51 & -- & 68.2 & -- \\
SFDG~\cite{2023dynamic} & 2023 &  & Freq; MD & EfficientNet~\cite{tan2019efficientnet} & FF++-LQ & unk. & 95.98 & 88.00 & 75.83 & 73.64 \\ 
MAT~\cite{zhao2021multi} & 2021 &  & Spatial; MTL, MD & EfficientNet~\cite{tan2019efficientnet} & FF++-HQ & Frame & 99.29 & -- & 67.44 & -- \\
Liang et al.~\cite{liang2022exploring} & 2022 &  & Spatial; MTL & ResNet~\cite{he2016deep} & FF++-DF-LQ & Frame & 99.22 & -- & 82.38 & -- \\
RECCE~\cite{cao2022end} & 2022 &  & Spatial; MTL, SSL & XceptionNet~\cite{chollet2017xception} & FF++-LQ & Frame & 95.02 & -- & 68.71 & 69.06 \\
Sun et al.~\cite{sun2021domain} & 2021 &  & Spatial; TL & EfficientNet~\cite{tan2019efficientnet} & FF++-HQ & Frame & 98.5 & -- & 64.1 & 69.0\\
FReTAL~\cite{kim2021fretal} & 2021 &  & Spatial; TL & XceptionNet~\cite{chollet2017xception} & FF++-HQ & Frame & 99.25 & -- & -- & --\\
Nadimpalli et al.~\cite{nadimpalli2022improving} & 2022 &  & Spatial; TL & EfficientNet~\cite{tan2019efficientnet} & FF++-HQ & Frame & 99.4 & -- & 66.9 & -- \\
Face-Cutout~\cite{das2021towards} & 2021 &  & Spatial; Aug & EfficientNet~\cite{tan2019efficientnet} & FF++-LQ & Frame & 98.77 & -- & -- & -- \\
RFM~\cite{wang2021representative} & 2021 &  & Spatial; Aug & XceptionNet~\cite{chollet2017xception} & CD2 & Frame & -- & -- & 99.97 & --\\
SIA~\cite{sun2022information} & 2022 &  & Spatial; MD & EfficientNet~\cite{tan2019efficientnet} & FF++-LQ & unk. & 93.45 & -- & 77.35 & -- \\
\hline
LipForensics~\cite{haliassos2021lips} & 2022 & \multirow{18}{*}{Video} & PC; FE, SSL & ResNet~\cite{he2016deep}, MS-TCN~\cite{martinez2020lipreading} & FF++-HQ & Video & 99.7 & -- & 82.4 & 73.5 \\
LRNet~\cite{sun2021improving} & 2021 & & PC; FE & RNN & FF++-HQ & Video & 97.3 & -- & 56.3 & -- \\
Masi et al.~\cite{masi2020two} & 2020 & & Freq; TL, MD & DenseNet~\cite{huang2017densely}, LSTM & FF++-LQ & Video & 91.10 & --& 73.41 & -- \\
CD-Net~\cite{song2022adaptive} & 2022 & & Freq, GS; TL, MD & XceptionNet~\cite{chollet2017xception}, SDM & FF++-HQ & Video & 99.9 & -- & 88.5 & 77.0 \\
RealForensics~\cite{haliassos2022leveraging} & 2022 & & AV; MTL, SSL & CSN~\cite{tran2019video}, ResNet~\cite{he2016deep} & FF++-HQ & Video & -- & -- & 86.9 & 75.9 \\
TD-3DCNN~\cite{zhang2021detecting} & 2021 & & GS; Aug & 3DCNN & FF++-unk. & Video & 72.22 & -- & 57.32 & 55.02 \\
FInfer~\cite{hu2022finfer} & 2022 & & GS; SSL & SDM, GRU~\cite{oord2018representation} & FF++-unk. & Video & 95.67 & -- & 70.60 & 70.39\\
STIL~\cite{gu2021spatiotemporal} & 2021 & & GS; MD & ResNet~\cite{he2016deep}, SDM & FF++-LQ & Video & 97.12 (DF) & -- & 75.58 & -- \\
RobustForensics~\cite{song2022face} & 2022 & & GS; MD & XceptionNet~\cite{chollet2017xception}, SDM & FF++-LQ & Video & 95.10 & -- & 79.0 & --\\
AltFreezing~\cite{wang2023altfreezing} & 2023 & & GS; MD & 3D ResNet50~\cite{carreira2017quo} & FF++-HQ & Video & 99.7 & 98.5 & 89.5 & -- \\
Zhang et al.~\cite{zhang2022deepfake} & 2022 & & LS; Aug & ViT~\cite{dosovitskiy2020image} & FF++-DF-raw & Video & 99.76 & -- & 69.78 & 66.99 \\
DIANet~\cite{ijcai2021p102} & 2021 & & LS; MD & ResNet~\cite{he2016deep}, SDM & FF++-HQ & Video & 98.8 & -- & 70.4 & -- \\
FTCN~\cite{zheng2021exploring} & 2021 & & LS; MD & 3D ResNet50~\cite{carreira2017quo}, ViT~\cite{dosovitskiy2020image} & FF++-HQ & Video & 99.7 & -- & 86.9 & 74.0 \\ 
LTTD~\cite{guan2022delving} & 2022 & & LS; MD & \makecell{3D CNN~\cite{hara2017learning}, ViT~\cite{dosovitskiy2020image}} & FF++-HQ & Video & 99.52 & -- & 89.3 & 80.4 \\
Gu et al.~\cite{gu2022delving} & 2022 & & GLT; MD & ResNet50~\cite{carreira2017quo}, SDM & FF++-LQ & Video & 98.19 (DF) & -- & 77.65 & 68.43 \\
\hline
\end{tabular}
}
\end{threeparttable}
\begin{threeparttable}
\begin{tablenotes}
    \tiny
      \item[*] PC: Physical Characteristics, GP: Generation Pipeline, Freq: Frequency, AV: Audio-visual, GS: Global-spatial, LS: Local-spatial, GLT: Global-and-local-temporal, FE: Feature Engineering, TTL: Multi-task Learning, TL: Transfer Learning, Aug: Data/Feature Augmentation, SSL: Self-supervised Learning, MD: Module Design, SDM: Self-designed Module.
      \item[*] FF++: FaceForensics++ dataset~\cite{rossler2019faceforensics++}, DFD:  DeepFakeDetection Dataset~\cite{dfd2019}, CD2: Celeb-DF-v2 dataset~\cite{li2020celeb}, DFDC: \cite{dolhansky2020deepfake}.
      \item[*] Video-level AUC is computed by averaging the classification scores of the video frames.
\end{tablenotes}
\end{threeparttable}
\end{table*}

\begin{table*}[!t]
\caption{Comparison of universal generative visual media detection methods. 
\label{tab:universal_detection_method_comparison}}
\centering
\begin{threeparttable}
      \resizebox{1\textwidth}{!}{
\begin{tabular}{l|c|c|c|c|c|c}
\hline
\multirow{2}{*}{Method} & \multirow{2}{*}{Year} & \multicolumn{2}{c|}{Taxonomy} & \multirow{2}{*}{Backbone} & \multirow{2}{*}{\makecell{Training\\Dataset}} & \multirow{2}{*}{(Test Dataset) Performance} \\
& & Modal & Subcategory & & &  \\
\hline
CNNDetection~\cite{wang2020cnn} & 2020 & \multirow{12}{*}{Image} & Spatial; DA & ResNet & CNNDetection & (CNNDetection) AP=93.0\\
HiFi-Net~\cite{guo2023hierarchical} & 2023 &  & GP; MTL, MD & HRNet & HiFi-IFDL & (HiFi-IFDL) AUC=96.8, F1=94.1; (CASIA) AUC=99.5, F1=97.4; (DFFD) AUC/PBCA=99.45/88.50 \\
FrePGAN~\cite{jeong2022frepgan} & 2022 & & Freq; MD & ResNet & CNNDetection (4 Classes) &  (CNNDetection) Acc=79.4, AP=87 \\
FingerprintNet~\cite{jeong2022fingerprintnet} & 2022 & & GP; SSL & ResNet  & CNNDetection (1 Class) &  (CNNDetection) Acc=82.6, AP=94.0\\
Liu et al.~\cite{liu2022detecting} & 2022 & & Freq, Spatial; FE, MD & ResNet & CNNDetection (20 Classes) &  (CNNDetection, HiSD, Glow) Acc=92.8, AP=94.8, F1=92.5\\
FatFormer~\cite{liu2024forgery} & 2024 & & Freq, Spatial; FE, MD& Transformer & CNNDetection (4 Classes) & (CNNDetection) Acc/AP=98.4/99.7, (UFD~\cite{ojha2023towards}, DIRE~\cite{wang2023dire}) Acc/AP=95.0/98.8\\
UnivFD~\cite{ojha2023towards} & 2023 & & Semantic; FE & Transformer & UFD & (UFD) Acc=81.38, AP=93.38\\
AEROBLADE~\cite{ricker2024aeroblade} & 2024 & & Spatial; FE & SD~\cite{rombach2022high} & Self-dataset & (Self-dataset) AP=99.71, TPR@5\%FPR=98.81\\
DE-FAKE~\cite{sha2023fake} & 2023 & & Semantic; FE & Transformer & Self-dataset & (Self-dataset) Acc=92.075, AUC=91.5, F1=86.475\\
NPR~\cite{tan2024rethinking} & 2024 & & GP; SSL & ResNet & CNNDetectin (20 Classes) & (CNNDetection) Acc/AP=92.5/96.1; (Self-dataset) Acc/AP=93.2/96.6; (DIRE) Acc/AP=95.3/99.8; (UFD) Acc/AP=95.2/97.4\\
LGrad~\cite{tan2023learning} & 2023 & & Spatial; FE & ResNet & ProGAN, CelebaHQ & (ProGAN-CelebaHQ) Acc/AP=96.4/99.5, (StyleGAN-CelebaHQ) Acc/AP=84.8/95.0, (StyleGAN2-CelebaHQ) Acc/AP=81.9/93.6\\
DIRE~\cite{wang2023dire} & 2023 & & Spatial; FE & ResNet & DIRE & (DIRE) Acc/AP=99.9/100.0 \\
\hline
DeMamba~\cite{chen2024demamba} & 2024 & \multirow{3}{*}{Video} & LP; MD & Mamba~\cite{gu2023mamba} & GenVideo & (GenVideo many-to-many) Acc/AP=94.42/97.10; (GenVideo one-to-many) Acc/AP=75.95/73.82\\
AIGVDet~\cite{bai2024ai} & 2024 & & Optical; MTL & GVD & ResNet & (Moonvalley) Acc/AUC=95.1/100; (VideoCraft) Acc/AUC=91.3/97.0; (Pika) Acc/AUC=89.5/95.5; (NeverEnds) Acc/AUC= 89.5/95.7\\
DuB3D~\cite{ji2024distinguish} & 2024 & & Optical; MD & GenVidDet & Swin-T~\cite{liu2022video} & (In-domain) Acc/F1=96.77/96.79; (Out-of-domain) Acc/F1=79.19/79.77\\
\hline
\end{tabular}
}
\end{threeparttable}
\begin{threeparttable}
\begin{tablenotes}
    \tiny
      \item[*] The abbreviations of subcategories in Taxonomy refer to Table~\ref{tab:deepfake_detection_method_comparison}.
\end{tablenotes}
\end{threeparttable}
\end{table*}

\begin{table*}[!t]
\caption{Comparison of disruption methods. The publication year, belonging categories based on the taxonomy introduced in Section~\ref{sec:disruption}, applicable scenario, target model, objective domain, evaluation dataset, and disruption performance of each method are summarized in the table.\label{tab:disruption_method_comparison}}
\centering
\begin{threeparttable}
      \resizebox{1\textwidth}{!}{
\begin{tabular}{l|c|c|c|c|c|c|c}
\hline
Method & Year & Taxonomy & Target Model & Scenario & Objective & Dataset & Performance \\
\hline
Ruiz et al.~\cite{ruiz2020disrupting} & 2020 & I; G & StarGAN~\cite{choi2018stargan} & Deepfake & Image & CelebA~\cite{liu2015deep} & $L_1=0.462$, $L_2=0.332$, \%dis=100\%\\
Huang et al.~\cite{huang2021initiative} & 2021 & I; G & Unet128 & Deepfake & Image & CelebA & DSR=100\%, $L_2=0.129$\\
CUMA-Watermark~\cite{huang2022cmua} & 2022 & I; G & StarGAN & Deepfake & Image & CelebA & $L^2_{\text{mask}}=0.20$, $SR_{\text{mask}}=100.0$, FID=201.003, ACS=0.286, TFHC=66.26\%$\rightarrow$20.61\%\\
ARS~\cite{guan2024adversarial} & 2024 & I; G & pSp-mix~\cite{richardson2021encoding} & Deepfake & Image & FFHQ~\cite{karras2019style} & MSE=1.3218, LPIPS=0.8007, SSIM=0.3602, PSNR=5.7422\\
LaS-GSA~\cite{yeh2021attack} & 2021 & I; G & CycleGAN~\cite{yeh2021attack} & Deepfake & Image & CelebA-HQ~\cite{karras2018progressive} & ASR=85\%, Q=42917\\
LAE~\cite{he2022defeating} & 2022 & I; G & StarGAN & Deepfake & Image & CelebA-HQ & $l_2=0.014$, MCD=0.10\\
\multirow{2}{*}{PhotoGuard~\cite{salman2023raising}} & \multirow{2}{*}{2023} & I; E & \multirow{2}{*}{SD v1.5} & \multirow{2}{*}{Editing} & Latent & \multirow{2}{*}{Self-dataset} & FID=130.6, PR=0.95, SSIM=0.58, PSNR=14.91, VIFp=0.30, FSIM=0.73\\
 & & I; G, D& & & Image & & FID=167.6, PR=0.87, SSIM=0.50, PSNR=13.58, VIFp=0.24, FSIM=0.69\\
Zhang et al.~\cite{zhang2023robustness} & 2023 & I; E, G, D & SD v1.5 & Variation & Latent & Coco~\cite{lin2014microsoft} & FID=206.3, SSIM=0.076, PSNR=11.82, CLIP=29.89\\
\multirow{2}{*}{Mist~\cite{liang2023mist}} & \multirow{2}{*}{2023} & \multirow{2}{*}{I; E, G} & SD + TI & \multirow{2}{*}{Personalization} & \multirow{2}{*}{\makecell{Latent, Noise}} & \multirow{2}{*}{WikiArt~\cite{wikiart}} & FID=454.39, precision=0.02\\
&  &  & SD + DB &  &  &  & FID=429.40, precision=0.04\\
AdvDM~\cite{liang2023adversarial} & 2023 & I; G & LDM + TI & Personalization & Noise & LSUN-CAT~\cite{yu2015lsun} & FID=127.04, precision=0.1708, recall=0.061\\
MFA-MVS~\cite{yu2024step} & 2024 & C; G & LDM & Inpainting & Noise & Places~\cite{zhou2017places} & FID=52.5, Delta E=12.77, PSNR=11.8, SSIM=0.40 \\
Xue et al.~\cite{xue2023toward} & 2024 & I; E, G & LDM & Personalization  & Latent, Noise & \makecell{anime, portrait,\\landscape, WikiArt} & \makecell{SSIM=0.698, PSNR=29.562, LPIPS=0.425, VRAM=8G,\\TIME=30s, P-Speed=0.25, HumanEval=4.55}\\
UDP~\cite{zhao2023unlearnable} & 2024 & I; G & DDPM & Generation & Noise & CIFAR-10~\cite{krizhevsky2009learning} & FID=112.67, Precision=23.67, Recall=10.47\\
\multirow{2}{*}{Anti-DreamBooth~\cite{van2023anti}} & \multirow{2}{*}{2023} & \multirow{2}{*}{I, G} & \multirow{2}{*}{SD v2.1+DB} & \multirow{2}{*}{Personalization} & \multirow{2}{*}{Noise} & VGGFace2~\cite{cao2018vggface2} & FDFR=0.63, ISM=0.33, SER-FQA=0.31, BRISQUE=36.42\\
& & & & & & CelebA-HQ & FDFR=0.31, ISM=0.50, SER-FQA=0.55, BRISQUE=38.57\\
\multirow{2}{*}{SimAC~\cite{wang2023simac}} & \multirow{2}{*}{2024} & \multirow{2}{*}{I; G} & \multirow{2}{*}{SD v2.1+DB} & \multirow{2}{*}{Personalization} & \multirow{2}{*}{Noise} & VGGFace2 & FDFR=80.27, ISM=0.31, SER-FQA=0.22, BRISQUE=40.71\\
& & & & & & CelebA-HQ & FDFR=87.07, ISM=0.31, SER-FQA=0.21, BRISQUE=38.86\\
\multirow{2}{*}{MetaCloak~\cite{liu2024metacloak}} & \multirow{2}{*}{2024} & \multirow{2}{*}{I; G} & \multirow{2}{*}{SD v2.1+DB} & \multirow{2}{*}{Personalization} & \multirow{2}{*}{Noise} & VGGFace2 & SDS=0.432, $\textrm{IMS}_{\textrm{CLIP}}=0.644$, $\textrm{IMS}_{\textrm{VGG}}=-0.151$, CLIP-IQAC=-0.440, LIQE=0.570\\
& & & & & & CelebA-HQ & SDS=0.305, $\textrm{IMS}_{\textrm{CLIP}}=0.608$, $\textrm{IMS}_{\textrm{VGG}}=-0.637$, CLIP-IQAC=-0.354, LIQE=0.438\\
DUAW~\cite{ye2023duaw} & 2024 & I; D & SD v2.1+DB & Personalization & Latent & WikiArt & CLIP-Score=0.6899, IL-NIQE=48.55 \\ 
CAAT~\cite{xu2024perturbing} & 2024 & I; G & SD v2.1 + DB & Personalization & Noise & CelebA-HQ & FR=0.64, FS=0.32, IR=-0.14, FID=371\\
ITA~\cite{zheng2023understanding} & 2024 & I; G & SD 1.5 + LoRA & Personalization & Noise & CelebA-HQ & CLIP-IQA=35.68\\
\hline
\end{tabular}}
\end{threeparttable}
\begin{threeparttable}
\begin{tablenotes}
    \tiny
      \item[*] I: Input, C: Condition, E: Encoder, G: Generator, D: Decoder.
      \item[*] SD: Stable Diffusion~\cite{rombach2022high}, DB: DreamBooth~\cite{ruiz2023dreambooth}, TI: Textual Inversion~\cite{gal2023an}.
\end{tablenotes}
\end{threeparttable}
\end{table*}

\section{Evaluation Methodology}\label{sec:evluation_methodology}
To comprehensively evaluate defense methods against AI-generated visual media, extensive evaluation methodologies are employed in existing literature. This section reviews the widely used evaluation datasets, criteria, and metrics. 

\subsection{Datasets}
Disruption and authentication methods based on proactive defense strategies are typically evaluated using real datasets (e.g. FFHQ~\cite{karras2019style, he2022defeating}, CelebA~\cite{liu2015deep, he2022defeating}, COCO~\cite{lin2014microsoft, wu2023sepmark}) in combination with different generative models. However, these datasets do not align closely with the primary focus of this survey—AI-generated visual media. Therefore, this section concentrates on datasets containing AI-generated data, specifically those used for detection. These datasets are briefly summarized according to the generation techniques discussed in Section~\ref{sec:Generation}.

Deepfake datasets comprise images or videos featuring both real and manipulated faces. Over time, these datasets have evolved in scale and diversity to reduce overfitting in detection models. Unseen test sets with complex perturbations have also been introduced to better reflect real-world inference conditions. Moreover, many datasets include multi-modality data and various annotations to support more sophisticated tasks. A summary is provided in Table~\ref{tab:deepfake_datasets}.

With the rapid advancements in generative models such as GANs and DMs, datasets containing diverse objects and scenes have been proposed to support more complex detection tasks. As summarized in Table~\ref{tab:universal_datasets}, most datasets are primarily used for benchmarking forgery detection methods. Exceptions include PAL4Inpaint~\cite{zhang2022perceptual} and PAL4VST~\cite{zhang2023perceptual}, which are designed specifically for perceptual artifact localization in inpainted images. In addition, HiFi-IFDL~\cite{guo2023hierarchical} and AutoSplice~\cite{jia2023autosplice} are widely used for evaluating forgery localization performance.

\subsection{Evaluation Criteria and Metrics}
This section summarizes the evaluation criteria and metrics used for both passive and proactive defense methods.

\subsubsection{Detection}
Detection is typically formulated as a classification task. Media-level forgery detection performs binary classification, while fine-grained forgery detection may involve multi-class classification. Therefore, standard machine learning evaluation metrics such as the area under the ROC curve (AUC), accuracy (ACC), and average precision (AP) are widely adopted. Based on these metrics, detection methods are generally evaluated under various scenarios to benchmark their performance comprehensively. This section reviews commonly used evaluation criteria and associated metrics.

\paragraph{Detection Ability}
Detection ability serves as the foundational evaluation criterion. It is typically evaluated under two scenarios: \textit{in-domain} and \textit{cross-domain} evaluation.

\textit{In-domain detection ability} evaluates models on test data drawn from the same distribution as the training data, including intra-manipulation~\cite{gu2022hierarchical, wang2020cnn} and intra-dataset evaluation~\cite{shiohara2022detecting, epstein2023online}. 

\textit{Cross-domain detection ability} measures the generalization ability of detection methods to unseen manipulations. Depending on the composition of the training and testing datasets, cross-domain evaluation can be categorized as 1-to-n~\cite{cao2022end, ojha2023towards}, n-to-1~\cite{sun2021domain, epstein2023online}, and n-to-m~\cite{gu2022hierarchical, zhuang2022uia} evaluations. 

Commonly used metrics for detection ability include AUC, ACC, AP, equal error rate (ERR), F1 score, and LogLoss~\cite{dolhansky2020deepfake, cao2022end}. Some metrics (e.g., ACC) are threshold-dependent, while others (e.g., AUC and AP) are threshold-less and ranking-based. Given that the base rate of real media is extremely larger than that of fake media in real-world distributions~\cite{dolhansky2020deepfake, wang2020cnn, das2021towards}, threshold-less metrics are generally preferred, as thresholding can introduce bias under class imbalance~\cite{thabtah2020data}. Besides, several specialized metrics~\cite{dolhansky2020deepfake, das2021towards} are also proposed for better evaluating forgery detection. 

\paragraph{Detection Robustness}
Detection ability is evaluated on high-quality visual media, while real-world scenarios such as online social networks frequently contain degraded or lossy content. Therefore, robustness to such degradations is a critical evaluation criterion for detection methods.

Detection robustness is generally evaluated on lossy media with various image-level and video-level post-processing operations. Common image-level perturbations~\cite{dolhansky2020deepfake, jiang2020deeperforensics, wang2020cnn, wang2023altfreezing} include JPEG compression, Gaussian blurring, and resizing, while video-level distortion is most often the video compression~\cite{jiang2020deeperforensics, rossler2019faceforensics++}. These perturbations are applied at varying strengths, from low to high levels. Accordingly, performance variation curves can be plotted to illustrate the degradation of detection performance under increasing levels of distortion~\cite{wang2020cnn, li2022blip, dong2022protecting, wang2023altfreezing}.

\paragraph{Qualitative Analysis}
Qualitative analysis commonly includes visualizations of saliency maps and feature spaces. Saliency maps are generated using techniques such as Grad-CAM~\cite{selvaraju2017grad} and Grad-CAM++\cite{chattopadhay2018grad}, highlighting the input regions the detection model focuses on when classifying samples~\cite{shiohara2022detecting, liu2024forgery}. Feature space visualization is commonly performed using t-SNE~\cite{van2008visualizing}, which shows the separation degree of feature vectors between samples of different classes~\cite{zhang2022deepfake, ojha2023towards}.

Based on the aforementioned evaluation criteria and metrics, the representative methods discussed in Section~\ref{sec:detection} are summarized in Table~\ref{tab:deepfake_detection_method_comparison} and Table~\ref{tab:universal_detection_method_comparison}.

\begin{table*}[!t]
\caption{Comparison of authentication methods. The publication year, belonging categories based on the taxonomy introduced in Section~\ref{sec:authentication}, scenario, generative model, experiment dataset, verification detection performance, and media quality performance of each method are summarized in the table.\label{tab:authentication_method_comparison}}
\centering
\begin{threeparttable}
      \resizebox{1\textwidth}{!}{
\begin{tabular}{l|c|c|c|c|c|c|c}
\hline
Method & Year & Taxonomy & Scenario & Generative Model & Dataset & Verification Performance & Quality Performance\\
\hline
FakeTagger~\cite{wang2021faketagger} & 2021 & I; B & Deepfake & DeepFaceLab & CelebA-HQ & Acc=0.968 & PSNR=32.45, SSIM=0.931 \\
DiffusionSheild~\cite{cui2023diffusionshield} & 2023 & I; S & I2I & DDMP & CIFAR10 & Acc=100.0 & LPIPS=0.0012 \\
FT-Shield~\cite{cui2025ft} & 2025 & I; S & Person. & SD + DB & WikiArt & TPR=99.5\%, FPR=0.18\% & FID=62.25 \\
Tree-Rings~\cite{wen2023tree} & 2023 & G; F & T2I & SD v2 & MS-COCO & AUC/T@1\%F=1.000/1.000 &  CLIP Score=0.364, FID=25.93\\
Stable Signature~\cite{fernandez2023stable} & 2023 & D; B & T2I & LDM & MS-COCO~\cite{lin2014microsoft} & Acc=0.99 & PSNR=30.0, SSIM=0.89, FID=19.6 \\
Xiong et al.~\cite{xiong2023flexible} & 2023 & D; B & T2I & LDM & LAION~\cite{schuhmann2021laion} & Acc=0.999 &  PSNR=32.51, SSIM=0.93, FID=6.35 \\
WOUAF~\cite{kim2024wouaf} & 2024 & D; B & T2I & SD v2-base & MS-COCO & Acc=0.99 & CLIP-score=0.73, FID=24.42, Time<1 sec \\
SepMark~\cite{wu2023sepmark} & 2023 & I; B & Deepfake & SimSwap, GANimation, StarGAN & CelebA-HQ & BER=48.98\% & PSNR=38.5646, SSIM=0.9328, LPIPS=0.0080\\
Asnani1 et al.~\cite{asnani2022proactive} & 2022 & I; S & Deepfake & StarGAN & CelebA-HQ & AP=100.0\% & PSNR=90\%+ \\
Zhao et al.~\cite{zhao2023proactive} & 2023 & I; B & Deepfake & StarGAN2 & CelebA-HQ & Acc/F1=0.85/0.87 & SSIM=0.94, PSNR=33.32 \\
AdvMark~\cite{wu2024watermarks} & 2024 & I; B & Deepfake & SimSwap, FOMM, StarGAN, StyleGAN & CelebA-HQ & Real/Fake Acc=97.32/79.37 & BER=0.029, PSNR=38.18, SSIM=0.928 \\
FaceSigns~\cite{neekhara2022facesigns} & 2022 & I; B & Deepfake & FaceSwap, SimSwap, FSFT & CelebA & BRA=51.61 & PSNR=35.43, SSIM=0.962 \\
\hline
\end{tabular}
}
\end{threeparttable}
\begin{threeparttable}
\begin{tablenotes}
    \tiny
      \item[*] I: Input, G: Generator, D: Decoder, B: Bit string, S: Spatial, F: Frequency.
      \item[*] T2I: Text-to-image generation, Person.: Personalization.
\end{tablenotes}
\end{threeparttable}
\end{table*}

\subsubsection{Disruption}
Disruption evaluation is conducted between the original generative outputs and the disrupted outputs. Commonly adopted \textit{disruption performance evaluation} metrics include $L_p$-norm distance~\cite{ruiz2020disrupting, he2022defeating, huang2021initiative}, Mean Square Error (MSE)~\cite{aneja2022tafim, guan2024adversarial}, Peak Signal-to-Noise Ratio (PSNR)~\cite{aneja2022tafim, guan2024adversarial}, Learned Perceptual Image Patch Similarity (LPIPS)~\cite{aneja2022tafim, guan2024adversarial}, and Success Rate (SR)~\cite{ruiz2020disrupting}. These metrics are calculated under both basic disruption settings and common media processing pipelines, where the latter evaluates the \textit{disruption robustness}. Additionally, transferable disruption performance across multiple generators and qualitative analysis are also commonly used evaluation measurements~\cite{aneja2022tafim, guan2024adversarial}. Based on these evaluation criteria and metrics, a comparison of representative disruption methods reviewed in Section~\ref{sec:disruption} is provided in Table~\ref{tab:disruption_method_comparison}. 

\subsubsection{Authentication}
The commonly adopted authentication evaluation criteria include \textit{verification performance evaluation} and \textit{media quality evaluation}. \textit{Verification performance evaluation} typically evaluates the detection and identification performance of watermarks, where the former is used for semi-fragile authentication and the latter for robust authentication. They can be measured using accuracy~\cite{fernandez2023stable}, AUC~\cite{wen2023tree, an2024benchmarking}, and TPR (True Positive Rate)~\cite{wen2023tree, cui2025ft} on both clean and post-processed watermarked media, where the post-processing is used for robustness evaluation. \textit{Media quality evaluation} evaluates the perceptual quality of watermarked images, typically measured by PSNR, Frechet Inception Distance (FID), and Structural Similarity Index (SSIM)~\cite{fernandez2023stable, wen2023tree, an2024benchmarking}. Based on these commonly used evaluation settings, a comparison of the representative authentication methods surveyed in Section~\ref{sec:authentication} is summarized in Table~\ref{tab:authentication_method_comparison}. 

\section{Recommendations for Future Work}\label{sec:recommendations_for_future_work}
This paper presents a comprehensive survey of defense methods and their trustworthiness against AI-generated visual media. Although significant progress has been achieved, several gaps remain that should be addressed in future research.
\subsection{Passive Defense}

\subsubsection{Greater Efforts in Fine-grained Forgery Detection}
Fine-grained forgery detection can enhance forgery detection and provide interpretability within a multi-task learning framework~\cite{chen2022self, masi2020two}. Moreover, specific fine-grained forgery detection tasks, such as sequential manipulation prediction~\cite{xia2024mmnet, shao2022detecting} and perceptual forgery localization~\cite{zhang2023perceptual}, can facilitate forgery recovery, thereby contributing to the improvement of generative models.

\subsubsection{Greater Efforts for Exploring Trustworthiness towards Diverse Detection Tasks}
The majority of research on the trustworthiness of detection primarily focuses on forgery detection. However, fine-grained forgery detection also has trustworthiness concerns~\cite{wu2024traceevader}, such as robustness and fairness issues. Since different detection tasks explore distinct forgery traces, future work should consider trustworthiness across various detection tasks.

\subsubsection{Improvement of Detection Benchmarks for Universal Generative Visual Media}
While comprehensive and fair benchmarks of deepfake detection are extensively explored~\cite{deng2024towards, yan2023deepfakebench}, a systematic benchmark for universal generative visual media detection remains lacking, considering most detection methods~\cite{wang2023dire, liu2022detecting, sha2023fake} construct new datasets for evaluations. This gap hinders comprehensive evaluations of detection ability, generalization ability, and robustness. Therefore, authoritative and widely recognized benchmarks are required. Furthermore, existing universal datasets lack publicly available extra test sets that include high-quality fake content generated by diverse models~\cite{ruiz2023dreambooth, li2025fractal} and degraded media for robustness testing. 

\subsubsection{Exploration of Comprehensive Evaluation Criteria}
Most current evaluation criteria focus on either qualitative or quantitative evaluations of defense performance. However, given the practical need to process large volumes of visual content on social platforms, additional practicality metrics, such as inference time, floating-point operations per second (FLOPs), and model parameter size, should also be considered as important evaluation indicators.

\subsection{Proactive Defense}
\subsubsection{Exploring Diverse Semi-fragile Authentication Tasks}
The authentication tasks corresponding to fine-grained forgery detection tasks introduced in Section \ref{sec:fine-grained_forgery_detection} can also be explored with a semi-fragile watermarking framework. Moreover, the robustness of semi-fragile authentication can be further explored as the damaged watermark may be reconstructed~\cite{li2014semi, sabel2021robustness}.

\subsubsection{Exploring Diverse Trustworthiness Issues in Proactive Defense}
In passive defense, focused trustworthiness concerns include robustness and fairness. Similarly, proactive defense should also address a broader range of trustworthiness issues. 

\section{Conclusion}\label{sec:conclusion}
This survey provides a comprehensive overview of research on both proactive and passive defenses against AI-generated visual media. It covers the mainstream defense strategies: detection, disruption, and authentication, as well as their associated trustworthiness concerns. We define and formulate these defense strategies within a unified framework and propose a novel taxonomy for each, based on methodological characteristics that are applicable across defense tasks. In addition, we review evaluation methodologies, including evaluation datasets, criteria, and metrics used to evaluate these defenses. Finally, we discuss recommendations for future research to highlight promising directions. We hope this survey provides researchers with an in-depth understanding of defenses against diverse types of deep generative visual forgeries and appeals to further contributions.


\begin{acks}
The authors would like to thank the anonymous reviewers for their insightful comments and valuable suggestions. This research is supported by the National Key Research and Development Program of China (2023YFB3107401), the National Natural Science Foundation of China (T2341003, 62376210, 62161160337, 62132011, U24B20185, U21B2018, 62206217, 62406240, U244120060), and the Shaanxi Province Key Industry Innovation Program (2023-ZDLGY-38). Thanks to the New Cornerstone Science Foundation and the Xplorer Prize.
\end{acks}

\bibliographystyle{ACM-Reference-Format}
\bibliography{egbib_abbr}

\end{document}